\documentclass[final, 12pt]{elsarticle}

\usepackage{amssymb}
\usepackage{amsmath}

\usepackage{graphicx}
\usepackage{subcaption}
\usepackage{geometry}
\usepackage{ragged2e} 
\usepackage{pifont} 
\usepackage{xcolor} 
\usepackage{soul}
\usepackage{amsmath}
\usepackage{url}
\usepackage{arydshln}
\usepackage{multirow}
\usepackage[most]{tcolorbox}
\tcbuselibrary{listingsutf8}
\definecolor{introcolor}{HTML}{0056B3}       
\definecolor{titlecolor}{HTML}{28A745}       
\definecolor{descriptioncolor}{HTML}{343A40} 
\definecolor{guidecolor}{HTML}{DC3545}       

\journal{Knowledge-Based Systems}

\begin{document}
\begin{frontmatter}



\title{Zero-shot narrative detection in social messaging}


\author[label1]{Jesús M. Fraile-Hernández} 
\ead{jfraile@lsi.uned.es}
\cortext[label1]{Corresponding author}
\author[label1]{Anselmo Peñas}
\affiliation[label1]{organization={UNED NLP \& IR Group, Universidad Nacional de 
  Educación a Distancia},
            city={Madrid},
            postcode={28040},
            country={Spain}
            }
\author[label2]{Patrick Giedemann}
\affiliation[label2]{organization={Zurich University of Applied Sciences},
             city={Winterthur},
             postcode={8400},
             state={Zurich},
             country={Switzerland}}

\begin{abstract}
This study investigates the zero-shot ability of large language models (LLMs) to identify and classify hidden narratives in social messages. Our research hypothesis is that LLMs' extensive contextual knowledge allows them to interpret messages on a deeper, pragmatic level, going beyond basic sentiment or topic analysis. Experiments on the Dipromats and SemEval datasets show that providing models with human-written narrative descriptions significantly improves performance, without the need of training examples. In contrast, automatically generated descriptions or the use of few examples (few-shot) often degrade accuracy due to subtle shifts in framing. The study also finds that ensemble methods, particularly majority voting, enhance robustness and that larger models perform best while also being less sensitive to prompt variations. The findings validate that LLMs can effectively detect strategic narratives in a zero-shot setting, and when combined with simple ensembling and human-written descriptions, they can rival supervised systems, offering a scalable solution for narrative detection, specially when there is no training data for the vast majority of domains.
\end{abstract}

\begin{graphicalabstract}
\centering
\includegraphics[width=0.75\linewidth]{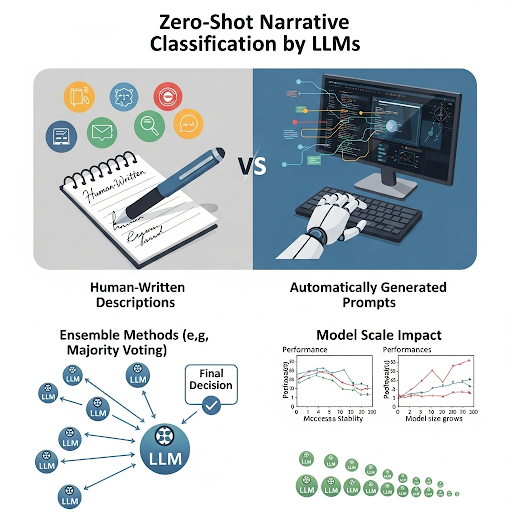}
\end{graphicalabstract}

\begin{highlights}
    \item First comprehensive zero-shot evaluation of narrative and subnarrative classification on the Dipromats and SemEval datasets.
    \item Systematic assessment of various prompting strategies (human-authored, automatically generated, and title-only).
    \item Evidence that human-curated descriptions consistently outperform other prompt types.
    \item Ensemble methods, particularly majority voting, significantly improve robustness and accuracy, especially for subnarrative detection.
    \item Larger models show superior performance and robustness, while some mid-sized models offer a good balance of stability and computational efficiency.
    \item Confirm the hypothesis that LLMs' extensive contextual knowledge allows them to interpret message communicative intentions in a zero-shot approach.
\end{highlights}


\end{frontmatter}


\section{Introduction}

The rapid advancements in LLMs have opened new frontiers in natural language processing, demonstrating impressive capabilities in tasks ranging from text generation to complex question answering. While much research has focused on their syntactic and semantic understanding, a critical, yet underexplored, area lies in measuring their capacity for pragmatic interpretation – specifically, their ability to discern deeper communicative intentions and the underlying narratives embedded within social messages.

Human communication is inherently complex, often extending beyond the literal meaning of words. As Sperber and Wilson's Relevance Theory posits \cite{sperber1986relevance}, utterances carry a presumption of optimal relevance, guiding the hearer to infer the speaker's intended meaning by balancing cognitive effort and contextual effects. This inferential process is crucial for understanding not just \textit{what} is said, but \textit{why} it is said, and what strategic goals the speaker aims to achieve.

\subsection{Problem statement}
In social discourse, messages frequently serve as vehicles for particular narratives – structured accounts designed to shape perceptions, influence attitudes, or motivate actions. These narratives are not always explicitly stated but are subtly conveyed through framing, emphasis, and the strategic selection of information.

A good example can be seen in political communication about a seemingly neutral topic. Consider the following statement:

\begin{center}
"A new government bill proposes a tax cut."
\end{center}

On its own, this is just a statement of fact. However, it can be used to advance very different narratives.

\begin{itemize}
    \item The "Economic Growth" Narrative. This narrative frames the tax cut as a way to benefit everyone.

\begin{itemize}
    \item Framing: The tax cut is presented as "stimulus" or a way to "put more money back in people's pockets."
    \item Emphasis: The message highlights the potential for new jobs, increased consumer spending, and a more robust economy.
    \item Selective information: It focuses on the benefits to small business owners and families, while downplaying any potential increase in the national debt or reduction in government services.
\end{itemize}
The underlying objective is to convince the public that the policy is a universally positive step for prosperity.

\item The "Social Inequality" Narrative. This narrative frames the same tax cut as a policy that primarily benefits a select few at the expense of the many.

\begin{itemize}
    \item Framing: The tax cut is called a "giveaway to the rich" or a "corporate handout."
    \item Emphasis: The message highlights how a small percentage of wealthy individuals or large corporations will see the largest financial gains.
    \item Selective information: It focuses on the potential cuts to public services like schools, healthcare, or infrastructure, which may be needed to offset the lost tax revenue.
\end{itemize}
The underlying objective is to convince the public that the policy is unfair and will exacerbate social and economic divisions.
\end{itemize}

In both cases, the \textbf{topic} is the same (a tax cut), but the \textbf{narrative} is the deeper, persuasive story that shapes how the audience perceives that topic.

\subsection{Research hypothesis}
Our hypothesis is that the vast contextual knowledge and sophisticated pattern recognition abilities acquired by LLMs during their training equip them with a latent capacity to interpret social messages at this profound pragmatic level. We propose that LLMs may be capable of detecting strategic communicative intentions and, crucially, identifying the specific narratives that messages are designed to promote. This goes beyond mere sentiment analysis or topic extraction, delving into the underlying persuasive or manipulative objectives of communication.

\subsection{Objectives}

The primary goal of this research is to empirically measure the extent to which LLMs can perform the task of identifying and classifying these embedded narratives. A key challenge, and a central focus of this investigation, is to assess this capability in a zero-shot setting, meaning without providing the models with explicit examples of narratives or their classifications during the evaluation phase. By exploring LLMs' inherent capacity for pragmatic inference and narrative detection, this study aims to shed light on their potential for advanced social intelligence and their applicability in domains requiring nuanced understanding of human communication, such as disinformation detection, social listening, and conflict resolution. This research will contribute to our understanding of LLM capabilities beyond surface-level linguistic processing, pushing the boundaries towards truly intelligent interpretation of human intent.

Grounded in the pragmatic nature of narrative communication, this study seeks to assess the degree to which LLMs exhibit latent competence in interpreting and classifying strategic communicative intentions embedded in social discourse. Rather than treating narratives as surface-level lexical or topical features, we approach them as implicit pragmatic constructs, often conveyed through framing, emphasis, and presupposition.

The core objective is to evaluate the zero-shot capabilities of LLMs in detecting such narratives—without reliance on annotated training data or task-specific fine-tuning. 

To this end, our investigation is guided by the following specific goals:

\begin{enumerate}
  
    \item Identify the boundaries of the zero-shot capabilities of LLMs in narrative detection. Understand whether these limitations stem from model architecture, instruction design or contextual ambiguity.
    
    \item Benchmark different zero-shot approaches even with existing systems that rely on supervised learning or domain-specific fine-tuning.
    
    \item Elucidate the trade-offs between performance and generalisability, and to determine whether the proposed method offers competitive accuracy while avoiding the substantial costs of data annotation and model retraining.
    
    \item Demonstrate the robustness of zero-shot approaches across multiple thematic domains.
\end{enumerate}

By pursuing these objectives, we aim to contribute to the discourse on narrative identification in low-resource settings by offering a practical and empirically grounded alternative to conventional supervised approaches.

\subsection{Research Questions}

To consistently evaluate the capacity of LLMs to perform zero-shot narrative detection, and to explore the design decisions that may influence their performance, this study is guided by the following research questions:

\begin{itemize}
    \item \textbf{RQ1:} Are concise narrative titles sufficient for accurate zero-shot detection, or do models benefit significantly from extended narrative descriptions?
    
    \item \textbf{RQ2:} Does the use of original human-created narrative descriptions produce better results than automatically rewritten or improved versions generated by the models themselves?
    \begin{itemize}
        \item \textbf{RQ2.1:} Among the strategies for prompt improvement (e.g., summarisation, elaboration, or stylistic rewriting), which yields the highest performance in narrative detection tasks?
    \end{itemize}

    \item \textbf{RQ3:} Does the narrative detection performance improve when using self-generated narrative descriptions, or is it preferable to rely on descriptions produced by more capable models, such as GPT systems?
  
    \item \textbf{RQ4:} How effective are ensemble strategies in improving the robustness and accuracy of zero-shot narrative detection?
    
    \item \textbf{RQ5:} How does the difference in model size affect performance in narrative identification?
    
    \item \textbf{RQ6:} Are the narrative descriptions generated by LLMs sufficiently coherent, semantically accurate, and pragmatically acceptable to serve as reliable inputs for narrative detection?

\end{itemize}

\subsection{Contributions}

This work makes the following contributions to the study of zero-shot narrative detection using LLMs:

\begin{itemize}
    \item A first comprehensive evaluation of zero-shot narrative classification across two multilingual and multidomain datasets, Dipromats and SemEval, addressing both narrative and subnarrative detection tasks.
    
    \item A systematic assessment of a wide range of prompting strategies, including original human-authored narrative descriptions, automatically generated prompts by various LLMs, and concise title-only inputs, analysing their impact on detection performance.
    
    \item Evidence that original human-curated narrative descriptions consistently outperform both title-only prompts and automatically generated alternatives, highlighting the importance of editorially consistent and contextually faithful inputs.
    
    \item An investigation of the relative effectiveness of self-generated versus GPT-generated descriptions, revealing no significant overall advantage.
    
    \item An evaluation of ensemble methods demonstrating that majority voting ensembles substantially improve robustness and accuracy, particularly in fine-grained subnarrative detection.
    
    \item An analysis of the influence of model scale, confirming that larger LLMs not only exhibit superior overall performance but also greater robustness to prompt variation, while some mid-sized models offer a competitive balance between stability and computational cost.
    
    \item An assessment of the semantic coherence and pragmatic acceptability of LLM-generated narrative descriptions, emphasising that their utility depends on alignment with the annotation framework to avoid divergence from ground-truth labelling.
\end{itemize}

\subsection{Structure of the Paper}
\begingroup
The remainder of this paper is organised as follows. Section~\ref{sec:related_work} reviews the related work. Section~\ref{sec:experimentation_settings} presents the experimental setting, detailing the datasets and evaluation metrics. Section~\ref{sec:methodology} describes the methodology, including narrative description generation, classifier models, classification schemes, prompting strategies, and ensemble protocols. Section~\ref{sec:results} reports the results, covering quantitative analyses, confidence intervals, and a qualitative error study. Section~\ref{sec:comparison_sota} compares our approach with the state of the art. Section~\ref{sec:generalisation_languages} presents an evaluation of how the proposed approach generalises to other languages. Section~\ref{sec:scalability} provides a scalability analysis, reporting computational statistics and efficiency considerations. Finally, Sections~\ref{sec:conclusion} and~\ref{sec:future_work} draw conclusions and outline directions for future research.
\endgroup
\section{Related Work}
\label{sec:related_work}
\subsection{Theoretical Foundations and Definitions of Narrative}
A narrative is a complex and multifaceted concept, with various definitions depending on the context in which it is used. It can refer to broad ideological framings, storytelling patterns, or structured sequences of events that shape public perception and discourse. In an effort to synthesize various formulations of the concept, Dennison \cite{dennisonNarrativesReviewConcepts2021} proposed a refined definition of narratives as ''selective depictions of reality across at least two points that can include one or more causal claims, and are generally generalizable and can be applied to multiple situations, as opposed to specific stories.''

Early theoretical conceptions of narrative identify core structural elements as essential to their definition. Adam \cite{adamTextesTypesPrototypes2001} defines narratives as prototypical sequences exhibiting thematic unity and a chronological progression of events involving characters. In a more formalized structure, Chatman \cite{chatmanStoryDiscourseNarrative2007} distinguishes between the story (the chain of events and their participants) and the discourse (the mode of expression, whether verbal, visual, or performative). Toolan \cite{toolanNarrativeCriticalLinguistic2012} also frames narratives as perceived sequences of interconnected, non-random events anchored in time and space, involving agents such as organizations, persons, and places.

A key contribution to structuralist narrative theory is Greimas's Actantial Model \cite{greimasStructuralSemanticsAttempt1983}, which proposes six core roles: Subject, Object, Sender, Receiver, Helper, and Opponent. This model underpins a number of computational frameworks by offering a semantic scaffold for representing narrative functions. Building upon these formal definitions, Piper \cite{piperNarrativeTheoryComputational2021} introduces a theoretical scheme for computational narrative identification, emphasizing elements such as events, temporal anchoring, and spatial framing. In addition, the study highlights how human perceptions of narrative deviate from formal definitions, with empirical findings suggesting that readers rely not only on textual features but also on cognitive responses. These discrepancies highlight the challenges models face when encoding the complexity of human narrative comprehension.

\subsection{Narrative Taxonomies and Analysis Frameworks}
In media analysis, narratives are often studied to understand how information is framed, how it spreads, and what underlying themes emerge from large-scale text corpora. Several examples of formulations have been provided in previous taxonomies and datasets. Kotseva et al. \cite{kotsevaTrendAnalysisCOVID192023} created a three-level narrative taxonomy on \mbox{COVID-19} and used it to classify and analyze trends over time. Li et al. \cite{liClassifyingCOVID19Vaccine2023} focused on a flat taxonomy of anti-vax narratives, while Hughes et al. \cite{hughesDevelopmentCodebookOnline2021} presented a taxonomy of typical anti-vax narratives, organized on several common tropes and rhetorical strategies. Coan et al. \cite{coanComputerassistedClassificationContrarian2021} presented a two-level taxonomy for common instances of climate change denial in short snippets. Amanatullah et al. \cite{TellUsHow} presented a flat taxonomy of common pro-Russian narratives found in alleged pro-Kremlin influence campaigns related to the war in Ukraine.

Further synthesizing the field, Santana \cite{santanaSurveyNarrativeExtraction2023a} surveys existing NLP methods for narrative extraction, covering the detection of core elements-events, agents, time, and space-across various textual domains. Importantly, the work underscores the challenges of annotation in this abstract task and links these structural elements to the temporal construction of narratives over time.

\subsection{Computational Approaches and Challenges in Narrative Detection}
Various models of narrative analysis have been proposed in computational settings, each focusing on different dimensions such as temporal ordering, coherence, and socio-pragmatic functions \cite{mishlerModelsNarrativeAnalysis1995}.

In the space of unsupervised narrative detection, Wildemann et al. \cite{wildemannAutomatedIdentificationCompeting} propose a multi-stage pipeline combining topic modeling, event detection, and event linking to automatically structure competing narratives in political discourse. Similarly, Elfes et al. \cite{elfesMappingNewsNarratives2024} utilize Greimas's Actantial Model to extract narrative roles using LLMs, encode them into embeddings, and project the results into a bidimensional space for clustering narrative structures without supervision.

On the supervised end, Levi et al. \cite{leviDetectingNarrativeElements2022} constructed a manually annotated dataset spanning domains such as economics, health, and immigration, and evaluated several transformer-based models as baselines for narrative classification. The reliance on annotated corpora in this work reveals the high cost and subjectivity involved in training data acquisition for this task. In addition, Haouari et al. \cite{haouariUKElectionNarrativesDatasetMisleading2025} introduce a new dataset for narrative identification in the context of UK elections. Their evaluation of GPT-4o \cite{openaiGPT4TechnicalReport2024} in zero-shot and low-shot environments, with and without descriptive narrative inputs, provides subtle empirical evidence of the influence of context-giving descriptions on model performance.

Recent progress in narrative detection has been supported by the introduction of new shared tasks that provide standardized benchmarks, such as SemEval 2025 Task 10 and Dipromats 2024 Task 2. In this context, Singh et al. \cite{singhGateNLPSemEval2025Task2025} present a supervised methodology involving synthetic data generation and fine-tuning LLMs using LoRA techniques \cite{huLoRALowRankAdaptation2021}. In contrast, Eljadiri et al. \cite{eljadiriTeamINSALyon2SemEval2025} adopt a zero-shot approach based on a multi-agent system incorporating GPT-4o and GPT-4o-mini, avoiding any training data altogether. Similarly, Fraile-Hernandez et al. \cite{fraile-hernandezUNEDTeamSemEval2025Task} develop a hybrid zero-shot classifier that leverages topic detection prior to narrative classification, relying solely on the narrative title rather than contextual descriptions.

\section{Experimentation setting}
\label{sec:experimentation_settings}
This section presents the experimental framework adopted in this study. We begin by describing the datasets employed for narrative classification, highlighting their structure. Subsequently, we detail the evaluation metrics used to assess model performance, taking into account the specificities of each dataset.
\subsection{Datasets}
Choosing the right evaluation datasets is important to ensure that our study is based on realistic and socially meaningful content. To test the ability of LLMs to detect narratives in a zero-shot setting, we needed datasets that provide clear narrative definitions and cover a range of geopolitical and thematic areas. It was also important that these datasets reflect the complexity of narrative structures and include diverse language use.

We conduct our experiments using two benchmark datasets specifically designed for narrative identification: \textit{Dipromats 2024 Task~2} \cite{fraile-hernandezAutomaticIdentificationNarratives2025} and \textit{SemEval-2025 Task~10 Subtask~2} \cite{semeval2025task10}. Both resources are publicly available and include narrative titles as well as short descriptions, documented in official reports and a supporting repository.\footnote{Supporting repository: \url{https://github.com/JesusFraile/Zero_shot_narrative_detection_in_social_messaging}}

\subsubsection*{Dipromats 2024 Task 2}

This dataset addresses narrative detection in the political domain and is constructed in two languages, English and Spanish. It comprises tweets published by official diplomatic accounts from four major geopolitical actors: China, Russia, the European Union, and the United States of America. Each tweet is annotated for the presence of none, one or several predefined narratives, resulting in a multi-label classification task. Each geopolitical region includes a maximum of six different narrative labels.

Annotations were carried out using both the narrative titles and brief narrative descriptions provided to human annotators. The test set includes 200 tweets per language and per region, totalling 1,600 instances. Notably, the organisers of this task provided a highly limited set of training and development examples: only 102 annotated tweets, uniformly distributed with approximately four examples per label. This limitation made efficient supervised fine-tuning of the original data unfeasible, leading to the use of zero-shot, few-shot or minimally supervised approaches. To illustrate the task more concretely, we provide below an example of a tweet together with some of the narratives it supports.

\begin{tcolorbox}[title=Dipromats Dataset: Example of Narrative Annotation, colback=gray!5!white, colframe=gray!50!black, sharp corners=south]
\footnotesize
\textbf{Example tweet:} \\
\textit{It is the \#US, not China that takes a predatory world view. China upholds the vision of a community with a shared future for mankind and a neighborhood diplomacy of amity, sincerity, mutual benefit, and inclusiveness.}

\vspace{0.3em}
\textbf{Supported Narratives:}
\begin{itemize}
    \item \textbf{CH1: The West is immoral, hostile and decadent.} Tweets depict the West, primarily the US, as immoral and hostile, positioning China as a victim of their reckless behavior.
    \item \textbf{CH2: China is a benevolent power.} This narrative highlights China's cooperative stance, emphasizing support for justice, international law, economic development in other nations, and pursuit of mutual benefits.
\end{itemize}
\end{tcolorbox}

\subsubsection*{SemEval-2025 Task 10 Subtask 2}

This dataset comes from a shared task focused on multilingual narrative detection in the domain of news media. Although it includes five languages (English, Bulgarian, Portuguese, Hindi, and Russian), our experiments are restricted to the English subset.

The subtask covers two thematic and policy-relevant areas: the Russia–Ukraine war and climate change. These topics are prolific sources of narrative-rich content across both traditional and social media, enabling the study of narrative propagation in high-impact global contexts. While climate change invokes narratives related to science, economics, and sustainability, the Russia–Ukraine war gives rise to politically and ideologically charged narratives.

Narrative annotation in this dataset follows a three-tiered hierarchical structure. The top level indicates the overarching topic (e.g., ‘Russia–Ukraine War’, ‘Climate Change’). The second level identifies the main narratives, capturing extensive interpretative frameworks, while the third level specifies sub-narratives, offering finer-grained distinctions. An \textit{Other} category exists at both the topic and sub-narrative levels to accommodate content that does not align with predefined labels.

This dataset includes a total of 10 main narratives and 46 sub-narratives (36 specific and 10 labelled \textit{Other}) for climate change, and 11 main narratives with 49 sub-narratives (38 specific and 11 labelled \textit{Other}) for the Russia–Ukraine war. To illustrate the task more concretely, we provide below an example of a tweet together with some of the narratives it supports.

\begin{tcolorbox}[title=SemEval Dataset: Example of Narrative Annotation, colback=gray!5!white, colframe=gray!50!black, sharp corners=south, breakable]
\footnotesize
\textbf{Example news:} \\
\begin{itshape}
Putin says what Russia needs to do to win special operation in Ukraine 
Russia will win the special operation in Ukraine if the society shows consolidation and composure to the enemy, President Vladimir Putin said during a visit to the Ulan-Ude Aviation Plant on March 14, Rossiya 24 TV channel said.

Russia is not improving its geopolitical position in Ukraine. Instead, Russia is fighting "for the survival of Russian statehood, for the future development of the country and our children."

"In order to bring peace and stability closer, we, of course, need to show the consolidation and composure of our society. When the enemy sees that our society is strong, internally braced up, consolidated, then, without any doubt we will come to reach what we are striving for — both success and victory," Putin said.

According to him, many of the current problems began after the collapse of the Soviet Union, when they tried to put pressure on Russia to "destabilise the internal political situation." "Hordes of international terrorists" new sent to the purpose to accomplish this goal, Putin said.

Afterwards, the West decided to start rehabilitating Nazism in Russia's neighbouring states, including in Ukraine.

Nevertheless, Putin continued, Russia had long tried to build partnerships with both Western countries and Ukraine. However, after 2014, when the West contributed to the coup in Ukraine, the state of affairs changed dramatically. It was then when they started exterminating those who advocated the development of normal relations with Russia, he said.

According to Putin, Russia was forced to launch the special operation to protect the population. Western countries were hoping to break Russia quickly, but they were wrong, he said adding that Russia managed to raise its economic sovereignty since 2022.

Subscribe to Pravda.Ru Telegram channel, Facebook, RSS!

The fighting in several directions in the Kursk region continues. According to the Russian side, the Ukrainian Armed Forces are redeploying to attack in a new area
\end{itshape}

\vspace{0.3em}
\textbf{Supported Narratives and Subnarratives:}
\begin{itemize}
    \item \textbf{Russia is the Victim.} Statements that portray Russia as being unfairly targeted or victimized. Look for narratives that depict Russia as suffering unjust consequences.
    \begin{itemize}
        \item \textbf{Russia actions in Ukraine are only self-defence.} Statements that justify Russia’s action solely as legitimate self-defence and not a deliberate action.
    \end{itemize}
    \item \textbf{Blaming the war on others rather than the invader.} Statements attributing responsibility or fault to entities other than Russia in the context of Russia’s invasion of Ukraine. Look for direct or implied statements that shift blame away from Russia. Consider who is being held responsible for negative events or situations.
     \begin{itemize}
        \item \textbf{The West are the aggressors.} Statements that shift the responsibility for the conflict and escalation to the Western block. Look for direct or implied statements that mention that this conflict was a direct consequence of actions taken by the West. Consider who is being held responsible for negative events or situations.
    \end{itemize}
    \item \textbf{Hidden plots by secret schemes of powerful groups.} Statements that suggest hidden plots or secretive actions by powerful groups related to the war. Look for narratives involving clandestine activities, secret agendas, or unproven allegations. Focus on claims that lack credible evidence and suggest hidden motives.
     \begin{itemize}
        \item \textbf{Other.} 
    \end{itemize}
    \item \textbf{Discrediting Ukraine.} Statements that undermine the legitimacy, actions, or intentions of Ukraine or Ukrainians as a nation. Look for direct or implied statements that attack some aspect of the Ukrainian society.
     \begin{itemize}
        \item \textbf{Ukraine is associated with nazism.} Accusations that Ukrainian society or government has ties to or sympathies with Nazi ideology, often referencing historical events or extremist groups. This can go with discrediting Ukrainian nation, but should be used with any mention or hint of sympathy or association with (neo-)Nazism, historical or not.
    \end{itemize}
\end{itemize}
\end{tcolorbox}

In contrast to Dipromats, this task provided a substantial number of training and development examples, 2,054 annotated documents in several languages, which allowed many participating systems to adopt supervised learning strategies. The test set in English comprises 101 news articles. 

As with Dipromats, narrative titles and descriptions are available both in the official task documentation and through the shared repository.\footnotemark[1]

Leaderboards for both tasks are publicly available, providing reference results from participating systems and facilitating reproducibility and comparative evaluation.\footnote{Dipromats-2024 leaderboard: \\\url{https://huggingface.co/spaces/NLP-UNED/dipromats2024-task2-leaderboard}}%
\footnote{SemEval-2025 Task 10 leaderboard:\\\url{https://propaganda.math.unipd.it/semeval2025task10/leaderboard.php}}
\subsection{Evaluation Metrics}

To assess model performance, we adopt the evaluation protocols defined in the original tasks for both datasets. Given the subjective and inherently interpretative nature of narrative classification, different evaluation strategies are applied in each case to account for ambiguity and annotation variability.

\subsubsection*{Dipromats 2024 Task 2}

In this dataset, the annotation scheme follows a \textit{ternary annotation, binary classification} framework (A3C2). Annotators could assign one of three possible labels to each tweet with respect to a given narrative: \textit{yes}, \textit{no}, or \textit{leaning}. The \textit{leaning} label was introduced to capture instances where the support for a narrative was subtle, implicit, or open to interpretation. Despite the ternary nature of the annotations, the prediction task is framed as a binary classification problem, where models are expected to output either \textit{yes} or \textit{no} for each narrative. Evaluation metrics differ based on how the \textit{leaning} cases are treated, as discussed below.

As a result, three evaluation metrics were proposed, each reflecting a different treatment of the \textit{leaning} cases:
\begin{itemize}
    \item \textbf{F1-strict}, in which \textit{leaning} is treated as negative (i.e., equivalent to \textit{no}).
    \item \textbf{F1-lenient}, in which \textit{leaning} is interpreted as matching the system prediction, whether it is \textit{yes} or \textit{no}, thereby treating it as a flexible agreement with either label.
    \item \textbf{F1-average}, computed as the mean of F1-strict and F1-lenient.
\end{itemize}

In this study, although both F1-strict and F1-lenient are computed for completeness, we adopt F1-average (hereafter, F1) as the primary metric for performance evaluation. This choice reflects a balanced treatment of the ambiguous \textit{leaning} cases and facilitates comparability across narrative labels.

Regarding annotation reliability, inter-annotator agreement was estimated using Cohen’s $\kappa$. The reported agreement was $0.8584$ for Spanish annotations and $0.8030$ for English. However, these values were obtained considering the \textit{leaning} label and therefore do not directly reflect agreement on binary narrative classification. In this context, the F1-lenient metric is considered a more representative indicator of the expected annotation consistency in narrative identification.

\subsubsection*{SemEval-2025 Task 10 Subtask 2}

For this dataset, the official evaluation metric is the sample-based F1 averaged across documents (hereafter, F1). This metric calculates the F1-score for the predicted narrative and sub-narrative labels per instance, then averages it over the entire test set.

Inter-annotator agreement was reported using Krippendorff’s $\alpha$, with a score of 0.449 for main narratives and 0.388 for sub-narratives. These values indicate moderate agreement among human annotators and highlight the challenge of reliably identifying narrative structures at different levels of abstraction.

\section{Methodology}
\label{sec:methodology}

This section outlines the methodological framework employed in this study for zero-shot narrative detection. The process is structured into five interrelated steps detailed in the following subsections:

\begin{enumerate}
    \item Narrative Description Generation. 
    \item Prompting Strategy Design.
    \item Classifier Model Selection.
    \item Classification Scheme Definition.
    \item Evaluation Protocol and Setup.
\end{enumerate}

\subsection{Narrative Description Generation}
\label{sec:description_generation}
Several LLMs have been used to generate refined and more informative \\ \mbox{narrative/subnarrative (n/sn)} descriptions. These models have been prompted with varying levels of contextual input in order to evaluate how information richness affects the results obtained in the narrative identification task.
Three input configurations have been defined:

\begin{itemize}
    \item \textbf{Title only (Gen\_Title)}: The LLM receives the original title of the narrative, as defined by the human annotators.
    \item \textbf{Title + Description (Gen\_Title\_Desc)}: Both the narrative title and the human-authored description are provided as input.
    \item \textbf{Title + Description + Feature Guide (Gen\_Title\_Desc\_Guide)}: In addition to the title and the human-authored description, the model is instructed to enrich the generated narrative description by explicitly incorporating a set of analytical dimensions. These include: social, political, and economic context; key actors involved; emotions the message seeks to elicit; inferred conclusions intended for the reader; and relevant textual features. These values are obtained directly from the LLM’s generative output and are not annotated or verified by humans.
    
\end{itemize}

All generation scripts, prompts, and configuration details are made publicly available through the accompanying project repository.

A diverse set of LLMs was selected for this task, spanning a range of model sizes and architectures, in order to assess the influence of model capacity and pretraining characteristics on description generation. The following models were used: \texttt{gpt-4o-mini} \cite{openaiGPT4TechnicalReport2024}, \texttt{calme-2.4-rys-78b}~\cite{panahiMaziyarPanahiCalme24rys78bHugging2023} (a fine-tuned version of \texttt{Qwen 2 78B}~\cite{qwenQwen25TechnicalReport2025}), quantized to 4 bits, \texttt{Gemma-3 12B it} \cite{teamGemma3Technical2025}, \texttt{Exaone-3.5 7.8B} \cite{researchEXAONE35Series2024}, and \texttt{Granite-3.3 2B} \cite{IbmgraniteGranite332bInstruct2025}. This model heterogeneity enables us to compare the performance of compact, efficient generators with larger, more expressive LLMs.

\subsection{Classifier Models}
\label{sec:classifier_models}

Given the abstract and high-level nature of the task, narrative classification in a zero-shot setting requires models that possess extensive world knowledge and advanced reasoning capabilities. For this reason, a selection of competitive LLMs has been employed to serve as classifiers: \texttt{calme-2.4-rys-78b} (4-bit quantized), \texttt{Gemma-3 12B IT}, \texttt{Exaone-3.5 7.8B}, and \texttt{Granite-3.3 2B}. 

These models have been chosen based on two principal criteria. First, they are open-access and fully executable locally on hardware with less than 100GB of VRAM, ensuring transparency and reproducibility of results. Second, they are among the highest-performing models in their respective parameter ranges on the Massive Multitask Language Understanding (MMLU-Pro) benchmark as of March 2025. This selection enables a comparative analysis of how model scale and architectural diversity affect performance on a task as abstract and linguistically subtle as narrative detection.

The variation in model sizes—from 2B to 78B parameters—also allows for an exploration of scaling effects within the zero-shot classification paradigm, when combined with different prompting and narrative representation strategies.

For all models and runs, a temperature of 0.75 and top‑k of 5 was consistently applied, ensuring uniform generation conditions across experiments. The maximum number of labels predicted by the LLMs was unrestricted, allowing each model to output as many labels as deemed appropriate.
\subsection{Classification Schemes}
\label{sec:classification_schemes}

Given the structural differences between the two datasets, we adopt distinct classification schemes for each case.

For the Dipromats dataset, each instance is explicitly associated with a specific geopolitical region. Each region has its own set of narrative categories, and thus a region-specific multi-label classification is performed. Owing to the flat nature of the taxonomy, a single-step classification approach is sufficient.

In contrast, the SemEval dataset introduces a more complex hierarchical structure spanning three levels: \textit{topics}, \textit{narratives}, and \textit{sub-narratives}. As the primary evaluation in this task focuses on the detection of sub-narratives—from which narratives and topics can be inferred—we propose and compare three distinct classification schemes:

\begin{itemize}
    \item \textbf{All-in-One:} Sub-narratives are identified in a single step, using as input the complete list of sub-narratives along with their corresponding descriptions. 

    \item \textbf{Three-Step Hierarchical:} Classification is carried out through a three-step process. First, the topic is identified using only the list of possible topics as input. Second, narratives are classified using the narratives associated with the topic and their descriptions as input. Finally, for each narrative selected in the previous step, sub-narratives are identified using as input only the sub-narratives and their corresponding descriptions. 

    \item \textbf{Hybrid:} This strategy combines the previous two. The topic is first identified using as input the list of topics. Subsequently, sub-narratives associated with that topic are identified using as input the sub-narratives and their descriptions filtered by the predicted topic. 
\end{itemize}

\begin{figure}
    \includegraphics[width=1\linewidth]{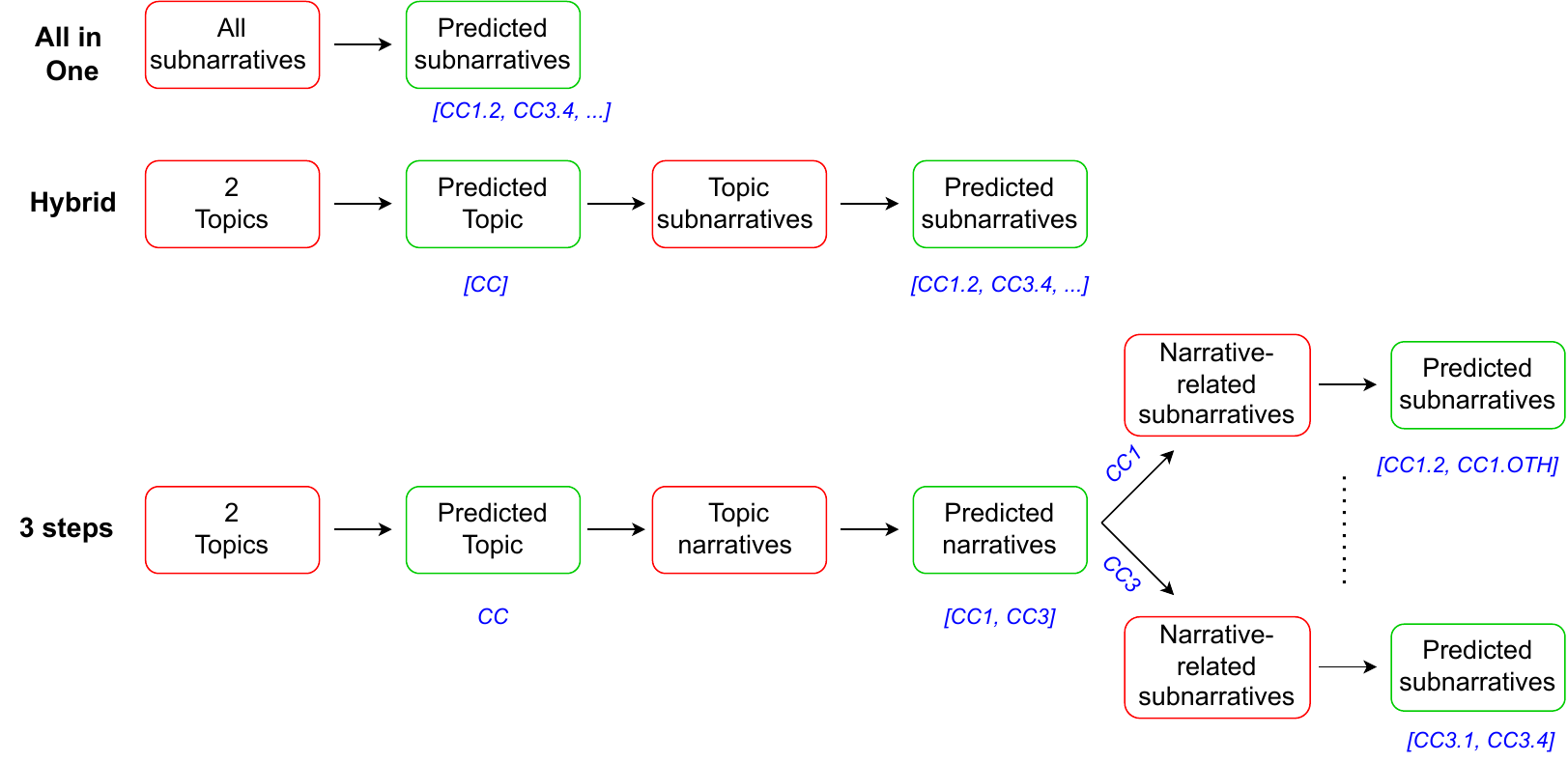}
    \caption{Overview of the three SemEval classification schemes.}
    \label{fig:sem_eval_classification_schemes}
\end{figure}

Figure \ref{fig:sem_eval_classification_schemes} illustrates the three classification strategies. Each red box denotes a model request, while green boxes represent the outputs obtained at each stage. Additionally, illustrative examples of these outputs are provided in blue to enhance interpretability. The All-in-One approach is the most efficient in terms of the number of requests, but at the cost of requiring extensive GPU memory and managing significant contextual load. The Three-Step scheme performs a lot more requests and may accumulate errors, but has the advantage that the context that is introduced in each request is smaller. Finally, the Hybrid scheme combines the advantages of both, accumulates fewer errors than the 3-step scheme by making only one first division, makes only 2 requests and has a much smaller context than the All in One scheme.

\subsection{Prompting Strategies}
\label{sec:prompting_strategies}

For each classifier model described in Section~\ref{sec:classifier_models}, we employed eight different prompting strategies that vary the (n/sn) descriptions used as input. These descriptions were generated according to the procedures outlined in Section~\ref{sec:description_generation}. To facilitate reference throughout this work, each prompting strategy is assigned a concise abbreviation reflecting its components and source of generation:

\begin{itemize}
    \item \textbf{Title}: Prompts include only the title of the (n/sn) without any accompanying description.
    \item \textbf{OrigDesc}: Prompts comprise the title alongside the original human-authored description.
    \item \textbf{Self\_Gen\_Title}: Prompts comprise the title plus a description generated by the classifier model itself based solely on the title (Gen\_Title).
    \item \textbf{GPT\_Gen\_Title}: Prompts include the title and a description generated by the GPT model from the title only.
    \item \textbf{Self\_Gen\_Title\_Desc}: Prompts include the title and a description generated by the classifier model itself from both the title and original description (Gen\_Title\_Desc).
    \item \textbf{GPT\_Gen\_Title\_Desc}: Prompts include the title and a description generated by the GPT model from both the title and original description.
    \item \textbf{Self\_Gen\_Title\_Desc\_Guide}: Prompts consist of the title and a description generated by the classifier model itself based on the title, original description, and additional feature guide prompts \\(Gen\_Title\_Desc\_Guide).
    \item \textbf{GPT\_Gen\_Title\_Desc\_Guide}: Prompts include the title and a description generated by the GPT model incorporating the title, original description, and the feature guide prompts.
\end{itemize}

By systematically applying this comprehensive suite of prompting strategies, we are able to rigorously evaluate the impact of varying levels of contextual and descriptive richness on zero-shot narrative detection performance. This approach allows us to dissect the contribution of each description type, from minimal only title prompts to richly guided narrative representations, and assess how different models respond to these input variations. 




The final prompt structure used across experiments follows the general design presented in Appendix A, with modifications adapted to each specific prompting strategy. All scripts used for model inference, including prompt definitions and hyperparameter configurations, are available in the public project repository accompanying this work.

\subsection{Evaluation Protocol and Ensemble Strategies}
\label{sec:evaluation_protocol}
To ensure the reliability and robustness of the results, we execute each combination of classifier model and prompting strategy across five independent runs. All runs are performed using identical hyperparameter configurations and input conditions. By averaging the evaluation metric across these five executions, we mitigate the effect of stochastic variance and obtain a more stable estimation of model performance. This averaged score is taken as the definitive performance value for each specific model–prompt pairing.

In addition to the individual evaluations, we incorporate two ensemble strategies across the five runs of each combination:

\begin{itemize}
    \item \textbf{Majority Voting Ensemble}: For each instance and each possible label, we consider its occurrence across the five runs. A label is included in the final prediction if it is predicted in at least three out of five runs (i.e., strict majority). This approach emphasises consensus and mitigates outlier decisions from individual runs.
    
    \item \textbf{Inclusion Ensemble}: A more inclusive approach, where all labels predicted in any of the five runs are aggregated to form the final prediction. This strategy prioritises coverage and recall, potentially at the cost of increased false positives.
\end{itemize}

These ensemble techniques enable us to analyse the extent to which prediction variance affects final outputs and to explore whether aggregation can yield more reliable zero-shot classifications. Furthermore, to assess the benefits of model diversity, we construct cross-model ensembles for the two prompting strategies that exhibit the best overall performance. Specifically, we combine predictions from the four different classifier models, each executed over five runs (resulting in ensembles based on $4 \times 5$ predictions per instance), and apply both majority voting and inclusion strategies. This design facilitates a comprehensive investigation into the trade-offs between model-scale diversity, prompt informativeness, and ensemble robustness.

\section{Results}
\label{sec:results}

This section reports the results obtained for the narrative detection task under a zero-shot setting across the two datasets: Dipromats and SemEval. Each result corresponds to a specific combination of classifier model and prompting strategy, as defined in Section~\ref{sec:methodology}, with performance computed as the average over five independent runs.

For the Dipromats dataset, Table~\ref{tab:dipromats_results} presents the macro F1 scores and std  for each model–prompt combination, broken down by language (Spanish and English) and classifier (Calme, Gemma, Exaone, Granite). The rows are grouped by the type of narrative input used: from original narrative titles (\texttt{Title}) and descriptions (\texttt{OrigDesc}) to different self-generated and GPT-generated narrative formulations. Each block includes the base classification result, calculated as the average macro F1 score across five independent runs. The subsequent two rows show the performance of ensemble methods using majority voting and inclusion based merging, respectively.

For the SemEval dataset, results are presented separately for each of the three classification schemes described in Section~\ref{sec:classification_schemes}. Specifically, Table~\ref{tab:semeval_all_in_one} reports the results for the All-in-One approach, Table~\ref{tab:semeval_three_step} for the Three-Step Hierarchical strategy, and Table~\ref{tab:semeval_hybrid} for the Hybrid variant. Each table includes the macro F1 scores averaged across five runs and std for every classifier–prompt pairing. Ensemble results using both aggregation strategies are also included. For each configuration, the first column reports the F1 score for narrative prediction, while the second column reports the F1 score for subnarrative detection.

In Table~\ref{tab:semeval_all_in_one}, classifier–prompt configurations marked with a red cross ($\textcolor{red}{\boldsymbol{\times}}$) indicate cases where inference could not be completed due to prompt length exceeding the 100~GB VRAM available on the target hardware. These entries are omitted from comparative analysis but are included for completeness.

\begin{table}[]
\footnotesize
\begin{tabular}{l|cc|cc|cc|cc}
& \multicolumn{1}{c}{\textbf{es}} & \multicolumn{1}{c|}{\textbf{en}} & \multicolumn{1}{c}{\textbf{es}} & \multicolumn{1}{c|}{\textbf{en}} & \multicolumn{1}{c}{\textbf{es}} & \multicolumn{1}{c|}{\textbf{en}} & \multicolumn{1}{c}{\textbf{es}} & \multicolumn{1}{c}{\textbf{en}} \\
& \multicolumn{2}{c|}{\textbf{Calme}}              & \multicolumn{2}{c|}{\textbf{Gemma}}              & \multicolumn{2}{c|}{\textbf{Exaone}}             & \multicolumn{2}{c}{\textbf{Granite}}            \\ \hline
\textbf{Title}                                                                    & 0.6345                 & 0.5834                 & 0.6125                 & 0.5647                 & 0.5519                 & 0.4990                 & 0.3336                 & \underline{0.3675}                 \\
std & 0.0032 & 0.0041 & 0.0030 & 0.0015 & 0.0083 & 0.0035 & 0.0139 & 0.0120 \\  

majority                                                                          & 0.6395                 & 0.5869                 & 0.6141                 & 0.5675                 & 0.5624                 & 0.5152                 & 0.3517                 & 0.3773                 \\
inclusion                                                                         & 0.6154                 & 0.5676                 & 0.6082                 & 0.5581                 & 0.5242                 & 0.4706                 & 0.3467                 & 0.3811                  \\ \hline
\textbf{OrigDesc}                                                                 & 0.6387                 & 0.6135                 & \underline{0.6242}                 & \underline{0.5932}                 & \underline{0.5725}                 & \underline{0.5348}                 & \underline{0.3352}                 & 0.3526                 \\
std & 0.0047 & 0.0023 & 0.0037 & 0.0020 & 0.0050 & 0.0053 & 0.0112 & 0.0056 \\  

majority                                                                          & 0.6396                 & 0.6150                 & 0.6266                 & 0.5937                 & 0.5895                 & 0.5496                 & 0.3351                 & 0.3590                 \\
inclusion                                                                         & 0.6216                 & 0.5934                 & 0.6251                 & 0.5855                 & 0.5488                 & 0.5056                 & 0.3557                 & 0.3683                 \\ \hline
\textbf{Self\_Gen\_Title}                                                         & 0.6210                 & 0.5714                 & 0.5762                 & 0.5313                 & 0.5062                 & 0.4837                 & 0.2861                 & 0.2984                 \\
std & 0.0048 & 0.0017 & 0.0025 & 0.0025 & 0.0050 & 0.0076 & 0.0102 & 0.0064 \\  

majority                                                                          & 0.6249                 & 0.5754                 & 0.5771                 & 0.5330                 & 0.5374                 & 0.5019                 & 0.2761                 & 0.2925                 \\
inclusion                                                                         & 0.5991                 & 0.5561                 & 0.5660                 & 0.5213                 & 0.4759                 & 0.4587                 & 0.3196                 & 0.3352                 \\ \hline
\textbf{\begin{tabular}[c]{@{}l@{}}Self\_Gen\_\\ Title\_Desc\end{tabular}}        & 0.6318                 & \underline{0.6182}                & 0.5864                 & 0.5699                 & 0.5463                 & 0.5120                 & 0.2889                 & 0.3295                 \\
std & 0.0062 & 0.0017 & 0.0020 & 0.0022 & 0.0093 & 0.0059 & 0.0102 & 0.0111 \\  

majority                                                                          & 0.6339                 & 0.6203                 & 0.5875                 & 0.5721                 & 0.5746                 & 0.5367                 & 0.2799                 & 0.3386                 \\
inclusion                                                                         & 0.6191                 & 0.6098                 & 0.5760                 & 0.5594                 & 0.5057                 & 0.4771                 & 0.3271                 & 0.3493                 \\ \hline
\textbf{\begin{tabular}[c]{@{}l@{}}Self\_Gen\_\\ Title\_Desc\_Guide\end{tabular}} & 0.6324                 & 0.6062                 & 0.5351                 & 0.4778                 & 0.4574                 & 0.4275                 & 0.2454                 & 0.2827                 \\

std & 0.0052 & 0.0060 & 0.0017 & 0.0029 & 0.0167 & 0.0089 & 0.0160 & 0.0036 \\  

majority                                                                          & 0.6322                 & 0.6084                 & 0.5341                 & 0.4757                 & 0.5002                 & 0.4554                 & 0.2175                 & 0.2760                 \\
inclusion                                                                         & 0.6237                 & 0.5939                 & 0.5292                 & 0.4839                 & 0.4208                 & 0.4067                 & 0.3046                 & 0.3228                 \\ \hline
\textbf{GPT\_Gen\_Title}                                                          & 0.6085                 & 0.5657                 & 0.5839                 & 0.5237                 & 0.5249                 & 0.4907                 & 0.2673                 & 0.2979                 \\
std & 0.0047 & 0.0049 & 0.0015 & 0.0027 & 0.0069 & 0.0055 & 0.0093 & 0.0093 \\  

majority                                                                          & 0.6111                 & 0.5729                 & 0.5827                 & 0.5242                 & 0.5585                 & 0.5177                 & 0.2569                 & 0.2872                 \\
inclusion                                                                         & 0.5928                 & 0.5462                 & 0.5758                 & 0.5162                 & 0.4844                 & 0.4629                 & 0.3218                 & 0.3327                 \\ \hline
\textbf{\begin{tabular}[c]{@{}l@{}}GPT\_Gen\_\\ Title\_Desc\end{tabular}}         & \underline{0.6403}                & 0.6135                 & 0.6138                 & 0.5787                 & 0.5458                 & 0.5243                 & 0.2848                 & 0.3008                 \\
std & 0.0047 & 0.0051 & 0.0021 & 0.0019 & 0.0061 & 0.0042 & 0.0107 & 0.0134 \\  

majority                                                                          & 0.6415                 & 0.6169                 & 0.6138                 & 0.5783                 & 0.5665                 & 0.5531                 & 0.2895                 & 0.2972                 \\
inclusion                                                                         & 0.6310                 & 0.6006                 & 0.6034                 & 0.5693                 & 0.5047                 & 0.4734                 & 0.3214                 & 0.3379                \\ \hline
\textbf{\begin{tabular}[c]{@{}l@{}}GPT\_Gen\_\\ Title\_Desc\_Guide\end{tabular}}  & 0.6266                 & 0.6048                 & 0.2747                 & 0.3802                 & 0.4846                 & 0.4378                 & 0.2206                 & 0.2717                 \\
std & 0.0054 & 0.0071 & 0.0060 & 0.0033 & 0.0078 & 0.0077 & 0.0068 & 0.0108 \\  

majority                                                                          & 0.6370                 & 0.6134                 & 0.2756                 & 0.3852                 & 0.5257                 & 0.4614                 & 0.2055                 & 0.2623                 \\
inclusion                                                                         & 0.6117                 & 0.5931                 & 0.3085                 & 0.3926                 & 0.4487                 & 0.4258                 & 0.2761                 & 0.3239                 
\end{tabular}
\caption{Dipromats results: macro F1 scores and std by model, prompting strategy, and language.}
\label{tab:dipromats_results}
\end{table}

\begin{table}[]
\footnotesize
\begin{tabular}{l|cc|cc|cc|cc}
& \multicolumn{2}{c|}{\textbf{Calme}}          & \multicolumn{2}{c|}{\textbf{Gemma}}          & \multicolumn{2}{c|}{\textbf{Exaone}}         & \multicolumn{2}{c}{\textbf{Granite}}        \\
\textbf{Title}                                                                    & 0.4296               & 0.2824               & 0.4004               & 0.2640               & \underline{0.3482}               & \underline{0.2128}               & \underline{0.1792}               & \underline{0.1202}               \\
std & 0.0227 & 0.0194 & 0.0059 & 0.0045 & 0.0072 & 0.0109 & 0.0168 & 0.0080 \\

majority                                                                          & 0.4420               & 0.2860               & 0.4010               & 0.2660               & 0.3720               & 0.2290               & 0.1970               & 0.1480               \\
inclusion                                                                         & 0.4560               & 0.3000               & 0.3960               & 0.2580               & 0.3510               & 0.2170               & 0.2120               & 0.1330               \\ \hline
\textbf{OrigDesc}                                                                 & \underline{0.4884}              & 0.3098               & \underline{0.4214}              & \underline{0.2862}               & 0.3434               & 0.2124               & 0.1666               & 0.1056               \\
std & 0.0161 & 0.0149 & 0.0079 & 0.0081 & 0.0174 & 0.0133 & 0.0234 & 0.0134 \\

majority                                                                          & 0.4980               & 0.3080               & 0.4360               & 0.3000               & 0.3600               & 0.2130               & 0.1960               & 0.1420               \\
inclusion                                                                         & 0.5080               & 0.3250               & 0.4230               & 0.2820               & 0.3740               & 0.2420               & 0.2030               & 0.1180              \\ \hline
\textbf{Self\_Gen\_Title}                                                         & 0.4362               & 0.3042               & 0.3392               & 0.2208               & 0.3016               & 0.1792               & 0.1424               & 0.0802               \\
std & 0.0231 & 0.0224 & 0.0068 & 0.0054 & 0.0172 & 0.0135 & 0.0142 & 0.0145 \\

majority                                                                          & 0.4490               & 0.3230               & 0.3460               & 0.2260               & 0.2950               & 0.1780               & 0.1500               & 0.1090               \\
inclusion                                                                         & 0.4990               & 0.3500               & 0.3440               & 0.2140               & 0.3570               & 0.2330               & 0.1900               & 0.1080                \\ \hline
\textbf{\begin{tabular}[c]{@{}l@{}}Self\_Gen\_\\ Title\_Desc\end{tabular}}        & 0.4852               & \underline{0.3286}               & 0.3592               & 0.2508               & 0.3106               & 0.1764               & 0.1246               & 0.0576               \\
std & 0.0237 & 0.0179 & 0.0150 & 0.0133 & 0.0226 & 0.0188 & 0.0172 & 0.0218 \\

majority                                                                          & 0.5160               & 0.3560               & 0.3560               & 0.2510               & 0.2770               & 0.1570               & 0.1740               & 0.1350               \\
inclusion                                                                         & 0.5040               & 0.3520               & 0.3740               & 0.2500               & 0.3990               & 0.2270               & 0.1740               & 0.0770                \\ \hline
\textbf{\begin{tabular}[c]{@{}l@{}}Self\_Gen\_\\ Title\_Desc\_Guide\end{tabular}} & 0.4522               & 0.3188               & 0.2810               & 0.2810               & 0.2836               & 0.1890               & 0.1028               & 0.0496               \\
std & 0.0184 & 0.0204 & 0.0055 & 0.0055 & 0.0241 & 0.0132 & 0.0073 & 0.0061 \\

majority                                                                          & 0.4550               & 0.3240               & 0.2770               & 0.2770               & 0.2760               & 0.1990               & 0.1780               & 0.1430               \\
inclusion                                                                         & 0.5080               & 0.3470               & 0.2820               & 0.2810               & 0.3600               & 0.2340               & 0.1610               & 0.0770                \\ \hline
\textbf{GPT\_Gen\_Title}                                                          & 0.4442               & 0.2810               & 0.3556               & 0.2350               & 0.2892               & 0.1600               & 0.1274               & 0.0496               \\
std & 0.0202 & 0.0200 & 0.0103 & 0.0083 & 0.0118 & 0.0199 & 0.0208 & 0.0054 \\

majority                                                                          & 0.4590               & 0.2810               & 0.3630               & 0.2370               & 0.2780               & 0.1500               & 0.1540               & 0.1130               \\
inclusion                                                                         & 0.4940               & 0.3330               & 0.3540               & 0.2270               & 0.3520               & 0.2120               & 0.1790               & 0.0780               \\ \hline
\textbf{\begin{tabular}[c]{@{}l@{}}GPT\_Gen\_\\ Title\_Desc\end{tabular}}         & 0.4684               & 0.3222               & 0.3830               & 0.2676               & 0.3124               & 0.1770               & 0.1178               & 0.0406               \\
std & 0.0248 & 0.0269 & 0.0186 & 0.0186 & 0.0228 & 0.0236 & 0.0107 & 0.0090 \\
majority                                                                          & 0.4750               & 0.3250               & 0.3760               & 0.2640               & 0.3280               & 0.1850               & 0.1860               & 0.1280               \\
inclusion                                                                         & 0.5170               & 0.3630               & 0.3830               & 0.2630               & 0.3550               & 0.2190               & 0.1650               & 0.0690                \\ \hline
\textbf{\begin{tabular}[c]{@{}l@{}}GPT\_Gen\_\\ Title\_Desc\_Guide\end{tabular}}  & 0.4530               & 0.3236               & 0.2978               & 0.2114               & 0.2526               & 0.1630               & 0.1008               & 0.0340               \\
std & 0.0210 & 0.0135 & 0.0189 & 0.0130 & 0.0146 & 0.0123 & 0.0137 & 0.0066 \\

majority                                                                          & 0.4540               & 0.3300               & 0.3070               & 0.2200               & 0.2700               & 0.1780               & 0.1090               & 0.0920               \\
inclusion                                                                         & 0.4910               & 0.3480               & 0.3230               & 0.2240               & 0.3180               & 0.2020               & 0.1570               & 0.0590               
\end{tabular}
\caption{SemEval – Three-Step classification results: macro F1 scores and std by model and prompting strategy.}
\label{tab:semeval_three_step}
\end{table}

\begin{table}[]
\footnotesize
\begin{tabular}{l|cc|cc|cc|cc}
& \multicolumn{2}{c|}{\textbf{Calme}} & \multicolumn{2}{c|}{\textbf{Gemma}} & \multicolumn{2}{c|}{\textbf{Exaone}} & \multicolumn{2}{c}{\textbf{Granite}} \\
\textbf{Title}                                                                    & \underline{0.5206}           & \underline{0.3216}           & 0.2898           & 0.2878           & \underline{0.3632}            & 0.1464           & \underline{0.1950}            & \underline{0.1172}           \\
std & 0.0145 & 0.0265 & 0.0061 & 0.0048 & 0.0108 & 0.0046 & 0.0114 & 0.0176 \\ 

majority                                                                          & 0.5320           & 0.3400           & 0.2950           & 0.2940           & 0.3570            & 0.1470           & 0.2690            & 0.2080           \\
inclusion                                                                         & 0.5380           & 0.3380           & 0.3070           & 0.2970           & 0.3830            & 0.1770           & 0.2490            & 0.1300           \\ \hline
\textbf{OrigDesc}                                                                 & 0.4858           & 0.3018           & \underline{0.4278}          & \underline{0.3242}           & 0.3398            & 0.1402           & 0.1770            & 0.0800           \\
std & 0.0231 & 0.0145 & 0.0275 & 0.0210 & 0.0106 & 0.0036 & 0.0123 & 0.0138 \\ 

majority                                                                          & 0.1050           & 0.0360           & 0.4220           & 0.3230           & 0.3590            & 0.1550           & 0.2070            & 0.1420           \\
inclusion                                                                         & 0.3400           & 0.1950           & 0.4730           & 0.3390           & 0.3390            & 0.1510           & 0.2300            & 0.1120           \\ \hline
\textbf{Self\_Gen\_Title}                                                         & 0.4516           & 0.2602           & 0.3786           & 0.2174           & 0.3014            & \underline{0.1844}          & 0.1542            & 0.0616           \\
std & 0.0165 & 0.0179 & 0.0139 & 0.0117 & 0.0338 & 0.0272 & 0.0098 & 0.0130 \\ 

majority                                                                          & 0.4420           & 0.2690           & 0.3720           & 0.2180           & 0.3230            & 0.2120           & 0.2600            & 0.2090           \\
inclusion                                                                         & 0.4960           & 0.2950           & 0.3890           & 0.2380           & 0.3770            & 0.2140           & 0.2170            & 0.0950          \\ \hline
\textbf{\begin{tabular}[c]{@{}l@{}}Self\_Gen\_\\ Title\_Desc\end{tabular}}        & 0.4682           & 0.2856           & 0.3714           & 0.2046           & 0.2814            & 0.1774           & 0.1878            & 0.0782           \\
std & 0.0085 & 0.0077 & 0.0086 & 0.0091 & 0.0261 & 0.0317 & 0.0322 & 0.0075 \\ 

majority                                                                          & 0.4790           & 0.3070           & 0.3830           & 0.2050           & 0.3230            & 0.2370           & 0.2470            & 0.1720           \\
inclusion                                                                         & 0.5020           & 0.3010           & 0.3810           & 0.2250           & 0.3610            & 0.1890           & 0.2430            & 0.1120           \\ \hline
\textbf{\begin{tabular}[c]{@{}l@{}}Self\_Gen\_\\ Title\_Desc\_Guide\end{tabular}} & 0.4536           & 0.2698           & 0.3854           & 0.1814           & 0.2652            & 0.0810           & 0.0890            & 0.0288           \\
std & 0.0248 & 0.0160 & 0.0078 & 0.0091 & 0.0194 & 0.0105 & 0.0179 & 0.0121 \\ 

majority                                                                          & 0.4670           & 0.2940           & 0.3990           & 0.1860           & 0.3160            & 0.1450           & 0.1710            & 0.1550           \\
inclusion                                                                         & 0.4850           & 0.2920           & 0.4060           & 0.2070           & 0.3140            & 0.1140           & 0.1820            & 0.0650            \\ \hline
\textbf{GPT\_Gen\_Title}                                                          & 0.4690           & 0.2654           & 0.3422           & 0.1804           & 0.2952            & 0.1818           & 0.1582            & 0.0654           \\
std & 0.0132 & 0.0138 & 0.0105 & 0.0089 & 0.0257 & 0.0261 & 0.0185 & 0.0121 \\ 

majority                                                                          & 0.4690           & 0.2850           & 0.3250           & 0.1600           & 0.3020            & 0.1960           & 0.2070            & 0.1620           \\
inclusion                                                                         & 0.5020           & 0.2830           & 0.3590           & 0.1960           & 0.3870            & 0.2260           & 0.2310            & 0.1060            \\ \hline
\textbf{\begin{tabular}[c]{@{}l@{}}GPT\_Gen\_\\ Title\_Desc\end{tabular}}         & 0.4574           & 0.2670           & 0.3716           & 0.2182           & 0.2834            & 0.1776           & 0.1758            & 0.0826           \\
std & 0.0096 & 0.0131 & 0.0158 & 0.0144 & 0.0378 & 0.0296 & 0.0185 & 0.0115 \\ 
majority                                                                          & 0.4740           & 0.2940           & 0.3940           & 0.2270           & 0.3330            & 0.2340           & 0.1950            & 0.1590           \\
inclusion                                                                         & 0.4910           & 0.2760           & 0.3870           & 0.2250           & 0.3650            & 0.2090           & 0.2390            & 0.1190          \\ \hline
\textbf{\begin{tabular}[c]{@{}l@{}}GPT\_Gen\_\\ Title\_Desc\_Guide\end{tabular}}  & 0.4500           & 0.2692           & 0.3804           & 0.1926           & 0.2804            & 0.1360           & 0.1236            & 0.0440           \\
std & 0.0198 & 0.0152 & 0.0138 & 0.0080 & 0.0299 & 0.0230 & 0.0133 & 0.0105 \\ 
majority                                                                          & 0.4520           & 0.2810           & 0.3840           & 0.1820           & 0.2680            & 0.1560           & 0.2550            & 0.2160           \\
inclusion                                                                         & 0.4650           & 0.2850           & 0.3830           & 0.2140           & 0.3610            & 0.1800           & 0.1850            & 0.0780          
\end{tabular}
\caption{SemEval – Hybrid classification results: macro F1 scores and std by model and prompting strategy.}
\label{tab:semeval_hybrid}
\end{table}

\begin{table}[]
\footnotesize
\begin{tabular}{l|cc|cc|cc|cc}
& \multicolumn{2}{c|}{\textbf{Calme}}                                             & \multicolumn{2}{c|}{\textbf{Gemma}}                                             & \multicolumn{2}{c|}{\textbf{Exaone}}                                            & \multicolumn{2}{c}{\textbf{Granite}} \\
\textbf{Title}                                                                    & \underline{0.4038}                                 & 0.2144                                 & 0.2856                                 & 0.2842                                 & 0.2604                                 & 0.1682                                 & 0.1712            & 0.1196           \\
std & 0.0132 & 0.0115 & 0.0050 & 0.0050 & 0.0186 & 0.0152 & 0.0410 & 0.0334 \\ 

majority                                                                          & 0.4310                                 & 0.2300                                 & 0.2870                                 & 0.2870                                 & 0.2340                                 & 0.1860                                 & 0.2210            & 0.1910           \\
inclusion                                                                         & 0.4640                                 & 0.2620                                 & 0.2930                                 & 0.2860                                 & 0.3430                                 & 0.1880                                 & 0.2320            & 0.1370           \\ \hline
\textbf{OrigDesc}                                                                 & 0.3806                                 & 0.2112                                 & \underline{0.2916}                                 & \underline{0.2844}                                 & \underline{0.2748}                                 & \underline{0.1872}                                 & 0.1846            & 0.1186           \\
std & 0.0234 & 0.0174 & 0.0117 & 0.0125 & 0.0216 & 0.0173 & 0.0375 & 0.0372 \\ 

majority                                                                          & 0.3820                                 & 0.2250                                 & 0.2770                                 & 0.2770                                 & 0.3440                                 & 0.3010                                 & 0.2770            & 0.2280           \\
inclusion                                                                         & 0.4420                                 & 0.2480                                 & 0.3340                                 & 0.3070                                 & 0.3290                                 & 0.1850                                 & 0.2240            & 0.1210         \\ \hline
\textbf{Self\_Gen\_Title}                                                         & 0.3578                                 & 0.1880                                 & 0.2704                                 & 0.2434                                 & 0.1898                                 & 0.1254                                 & 0.1968            & 0.1612           \\
std & 0.0131 & 0.0078 & 0.0343 & 0.0314 & 0.0271 & 0.0172 & 0.0356 & 0.0338 \\ 

majority                                                                          & 0.3390                                 & 0.1800                                 & 0.2750                                 & 0.2620                                 & 0.1570                                 & 0.1210                                 & 0.2970            & 0.2870           \\
inclusion                                                                         & 0.4080                                 & 0.2350                                 & 0.3360                                 & 0.2610                                 & 0.2400                                 & 0.1340                                 & 0.2280            & 0.1430           \\ \hline
\textbf{\begin{tabular}[c]{@{}l@{}}Self\_Gen\_\\ Title\_Desc\end{tabular}}        & $\textcolor{red}{\boldsymbol{\times}}$ & $\textcolor{red}{\boldsymbol{\times}}$ & 0.2542                                 & 0.2420                                 & 0.1898                                 & 0.1188                                 & 0.1724            & 0.1396           \\
std & $\textcolor{red}{\boldsymbol{\times}}$ & $\textcolor{red}{\boldsymbol{\times}}$ & 0.0166 & 0.0185 & 0.0284 & 0.0225 & 0.0321 & 0.0315 \\ 

majority                                                                          & $\textcolor{red}{\boldsymbol{\times}}$ & $\textcolor{red}{\boldsymbol{\times}}$ & 0.2570                                 & 0.2570                                 & 0.2330                                 & 0.2050                                 & 0.2710            & 0.2570           \\
inclusion                                                                         & $\textcolor{red}{\boldsymbol{\times}}$ & $\textcolor{red}{\boldsymbol{\times}}$ & 0.2900                                 & 0.2340                                 & 0.2410                                 & 0.1180                                 & 0.2160            & 0.1260            \\ \hline
\textbf{\begin{tabular}[c]{@{}l@{}}Self\_Gen\_\\ Title\_Desc\_Guide\end{tabular}} & $\textcolor{red}{\boldsymbol{\times}}$ & $\textcolor{red}{\boldsymbol{\times}}$ & $\textcolor{red}{\boldsymbol{\times}}$ & $\textcolor{red}{\boldsymbol{\times}}$ & $\textcolor{red}{\boldsymbol{\times}}$ & $\textcolor{red}{\boldsymbol{\times}}$ & 0.1444            & 0.1102           \\
std & $\textcolor{red}{\boldsymbol{\times}}$ & $\textcolor{red}{\boldsymbol{\times}}$ & $\textcolor{red}{\boldsymbol{\times}}$ & $\textcolor{red}{\boldsymbol{\times}}$ & $\textcolor{red}{\boldsymbol{\times}}$ & $\textcolor{red}{\boldsymbol{\times}}$ & 0.0248 & 0.0266 \\

majority                                                                          & $\textcolor{red}{\boldsymbol{\times}}$ & $\textcolor{red}{\boldsymbol{\times}}$ & $\textcolor{red}{\boldsymbol{\times}}$ & $\textcolor{red}{\boldsymbol{\times}}$ & $\textcolor{red}{\boldsymbol{\times}}$ & $\textcolor{red}{\boldsymbol{\times}}$ & 0.2110            & 0.2080           \\
inclusion                                                                         & $\textcolor{red}{\boldsymbol{\times}}$ & $\textcolor{red}{\boldsymbol{\times}}$ & $\textcolor{red}{\boldsymbol{\times}}$ & $\textcolor{red}{\boldsymbol{\times}}$ & $\textcolor{red}{\boldsymbol{\times}}$ & $\textcolor{red}{\boldsymbol{\times}}$ & 0.1980            & 0.1170           \\ \hline
\textbf{GPT\_Gen\_Title}                                                          & 0.3814                                 & 0.2058                                 & 0.2782                                 & 0.2544                                 & 0.1596                                 & 0.0998                                 & 0.1936            & 0.1488           \\
std & 0.0196 & 0.0270 & 0.0168 & 0.0114 & 0.0181 & 0.0203 & 0.0346 & 0.0306 \\

majority                                                                          & 0.3680                                 & 0.2190                                 & 0.2770                                 & 0.2690                                 & 0.1450                                 & 0.1100                                 & 0.2720            & 0.2570           \\
inclusion                                                                         & 0.4290                                 & 0.2440                                 & 0.3090                                 & 0.2340                                 & 0.2120                                 & 0.1120                                 & 0.2420            & 0.1430          \\ \hline
\textbf{\begin{tabular}[c]{@{}l@{}}GPT\_Gen\_\\ Title\_Desc\end{tabular}}         & 0.3720                                 & \underline{0.2172}                                 & 0.2652                                 & 0.2544                                 & 0.1788                                 & 0.0958                                 & 0.1888            & 0.1394           \\
std & 0.0181 & 0.0136 & 0.0128 & 0.0180 & 0.0260 & 0.0249 & 0.0298 & 0.0200 \\
majority                                                                          & 0.4040                                 & 0.2450                                 & 0.2760                                 & 0.2690                                 & 0.1650                                 & 0.1340                                 & 0.2800            & 0.2480           \\
inclusion                                                                         & 0.4270                                 & 0.2460                                 & 0.3180                                 & 0.2720                                 & 0.2270                                 & 0.0930                                 & 0.2170            & 0.1250           \\ \hline
\textbf{\begin{tabular}[c]{@{}l@{}}GPT\_Gen\_\\ Title\_Desc\_Guide\end{tabular}}  & $\textcolor{red}{\boldsymbol{\times}}$ & $\textcolor{red}{\boldsymbol{\times}}$ & $\textcolor{red}{\boldsymbol{\times}}$ & $\textcolor{red}{\boldsymbol{\times}}$ & $\textcolor{red}{\boldsymbol{\times}}$ & $\textcolor{red}{\boldsymbol{\times}}$ & \underline{0.2008}            & \underline{0.1686}           \\
std & $\textcolor{red}{\boldsymbol{\times}}$ & $\textcolor{red}{\boldsymbol{\times}}$ & $\textcolor{red}{\boldsymbol{\times}}$ & $\textcolor{red}{\boldsymbol{\times}}$ & $\textcolor{red}{\boldsymbol{\times}}$ & $\textcolor{red}{\boldsymbol{\times}}$ & 0.0343 & 0.0299 \\

majority                                                                          & $\textcolor{red}{\boldsymbol{\times}}$ & $\textcolor{red}{\boldsymbol{\times}}$ & $\textcolor{red}{\boldsymbol{\times}}$ & $\textcolor{red}{\boldsymbol{\times}}$ & $\textcolor{red}{\boldsymbol{\times}}$ & $\textcolor{red}{\boldsymbol{\times}}$ & 0.2920            & 0.2870           \\
inclusion                                                                         & $\textcolor{red}{\boldsymbol{\times}}$ & $\textcolor{red}{\boldsymbol{\times}}$ & $\textcolor{red}{\boldsymbol{\times}}$ & $\textcolor{red}{\boldsymbol{\times}}$ & $\textcolor{red}{\boldsymbol{\times}}$ & $\textcolor{red}{\boldsymbol{\times}}$ & 0.2350            & 0.1480                                     
\end{tabular}
\caption{SemEval – All-in-One classification results: macro F1 scores and std by model and prompting strategy.}
\label{tab:semeval_all_in_one}
\end{table}

Table~\ref{tab:ensemble_best_prompts} presents the ensemble results obtained using the two best-performing prompting strategies overall: \texttt{Title} and \texttt{OriginalDesc}. For each strategy, we aggregate predictions across all five runs from each classifier model. The column labelled \textit{All} corresponds to the ensemble that combines outputs from all four classifier models, resulting in a total of $4 \times 5 = 20$ runs per instance. In contrast, the column \textit{No Granite} reports ensemble results excluding the \texttt{Granite-3.3 2B} model. This exclusion is motivated by the substantial architectural and capacity differences of \texttt{Granite}, a considerably smaller LLM compared to the other three models. In the case of majority-voting ensembles, labels are included if they appear in at least 11 out of 20 predictions when using all models (\textit{All}), or in at least 8 out of 15 predictions when using all models except Granite (\textit{No Granite}).

\begin{table}[]
\begin{tabular}{lc|cccc}
                          &      & \textbf{All}    & \textbf{All}       & \textbf{No Granite} & \textbf{No Granite} \\ \hline
                          &      & \textbf{F1 es}     & \textbf{F1 en}        & \textbf{F1 es}         & \textbf{F1 en}         \\ \hline
\textbf{Dipro-Title}      & maj. & \underline{0.6428 }         & 0.5977             & 0.6414              & 0.5869              \\
\textbf{}                 & inc. & 0.4001          & 0.3937             & 0.5176              & 0.4662              \\ \hdashline
\textbf{Dipro-OrigDesc}   & maj. & 0.6341          & 0.6046             & 0.6417              & \underline{0.6216}              \\
\textbf{}                 & inc. & 0.4048          & 0.3975             & 0.5375              & 0.4960              \\ \hline
\textbf{}                 &      & \textbf{F1 nar} & \textbf{F1 subnar} & \textbf{F1 nar}     & \textbf{F1 subnar}  \\ \hline
\textbf{SemEval-Title}    & maj. & 0.2870          & 0.2870             & 0.3000              & 0.3000              \\
\textbf{(all\_in\_one)}   & inc. & 0.3280          & 0.1930             & 0.3910              & 0.2410              \\ \hdashline
\textbf{SemEval-Title}    & maj. & 0.2900          & 0.2890             & 0.3120              & 0.2970              \\
\textbf{(hybrid)}         & inc. & 0.3270          & 0.1770             & 0.3800              & 0.2310              \\ \hdashline
\textbf{SemEval-Title}    & maj. & 0.4890          & 0.3780             & 0.4470              & 0.2930              \\
\textbf{(3\_steps)}       & inc. & 0.3100          & 0.1990             & 0.3960              & 0.2610              \\ \hdashline
\textbf{SemEval-OrigDesc} & maj. & 0.2990          & 0.2960             & 0.3280              & 0.3260              \\
\textbf{(all\_in\_one)}   & inc. & 0.3200          & 0.1840             & 0.3720              & 0.2320              \\ \hdashline
\textbf{SemEval-OrigDesc} & maj. & 0.2280          & 0.2280             & 0.1490              & 0.1490              \\
\textbf{(hybrid)}         & inc. & 0.2640          & 0.1380             & 0.2810              & 0.1610              \\ \hdashline
\textbf{SemEval-OrigDesc} & maj. & 0.5090          & \underline{0.4000}            & \underline{0.5560}              & 0.3900              \\
\textbf{(3\_steps)}       & inc. & 0.3030          & 0.1960             & 0.4170              & 0.2800             
\end{tabular}
\caption{Ensemble performance for the \texttt{Title} and \texttt{OriginalDesc} prompting strategies, using all classifier models (\textit{All}) and excluding \texttt{Granite} (\textit{No Granite}).}
\label{tab:ensemble_best_prompts}
\end{table}

\subsection{Results Analysis}
\label{subsec:results_analysis}

To facilitate the interpretation of the extensive set of results, this section is structured around the six research questions posed in the study. Unless otherwise indicated, values reported for the Dipromats dataset represent the average of F1 scores across Spanish and English, and for SemEval, across both narrative and subnarrative levels. All detailed calculations are available in a publicly shared spreadsheet within the supplementary repository.

\textbf{RQ1: Are concise narrative titles sufficient for accurate zero-shot detection, or do models benefit significantly from extended narrative descriptions?}

To address this research question, we analyse the performance differences between each prompting strategy and the baseline configuration using only narrative titles. For the Dipromats dataset, we compute these differences across all classifier–prompt combinations in both Spanish and English settings, as presented in Table~\ref{tab:dipromats_results}. For the SemEval dataset, we perform the same comparison across the two classification targets (narratives and sub-narratives) and for each classification scheme individually (Tables~\ref{tab:semeval_all_in_one},~\ref{tab:semeval_three_step}, and~\ref{tab:semeval_hybrid}). Specifically, for each configuration, we calculate the delta in macro F1 score between a given prompting approach and the Title baseline, and then average these deltas to obtain a general performance estimate for each strategy.

In Dipromats, employing original human-authored descriptions (OrigDesc) leads to an average performance improvement of 1.5 pp (percentage points) over using only narrative titles (Title). In contrast, using generated descriptions results in a mean decrease of approximately 4 pp. A more granular analysis reveals that the Self\_Gen\_Title and GPT\_Gen\_Title strategies yield a 3.4 pp drop, \\ Self\_Gen\_Title\_Desc and GPT\_Gen\_Title\_Desc yield a 0.7 pp drop, and \\Self\_Gen\_Title\_Desc\_Guide and GPT\_Gen\_Title\_Desc\_Guide result in more substantial degradations of 6 pp and 10.6 pp, respectively.

SemEval exhibits a similar trend. Using original descriptions leads to a modest 0.66 pp improvement over using titles alone. However, generated descriptions underperform by an average of 2.7 pp. Specifically, Self\_Gen\_Title, GPT\_Gen\_Title, Self\_Gen\_Title\_Desc, and GPT\_Gen\_Title\_Desc each produce an average decline of 2.6 pp, with negligible differences between them. The more detailed prompt variants, Self\_Gen\_Title\_Desc\_Guide and GPT\_Gen\_Title\_Desc\_Guide, result in 4.8 pp reductions, excluding the All-in-One setting which only the smallest model could process.

Overall, 28.44\% of generated approaches outperform title-only prompts, while 71.56\% do not. Comparing only Title vs. OrigDesc, the latter proves superior in 56\% of cases. When contrasting OrigDesc with generated descriptions, OrigDesc performs better in 87.1\% of configurations.

\textit{Conclusion:} While titles alone offer a reasonable baseline, original human authored descriptions consistently enhance performance. Automatically generated content, particularly when verbose, may introduce noise that offsets potential contextual gains.

\textbf{RQ2: Does the use of original human-created narrative descriptions produce better results than automatically rewritten or improved versions generated by the models themselves?}

Building on the approach used in RQ1, this analysis focuses on the difference in performance between each generated prompt configuration and the OrigDesc baseline. By computing and averaging these deltas across the same model and dataset settings, we assess whether automatically rewritten descriptions offer any consistent advantage over original human-authored texts.

As discussed above, OrigDesc yields the highest performance overall. In \mbox{Dipromats}, prompt strategies involving generated descriptions result in a 5.6 pp average decline. When using full descriptions (Title\_Desc and Title\_Desc\_Guide), performance drops by 7.5 pp and 12 pp, respectively. In SemEval, the decrease is 4.1 pp for narrative detection and 2.7 pp for subnarrative detection.

Only 12.9\% of all evaluated prompt configurations outperform the OrigDesc baseline.

Additionally, in deployment‑faithful scenarios where only titles or noisy/mismatched descriptions are available, model performance is expected to decrease. Our experiments show that manually curated narrative descriptions consistently outperform automatically generated prompts. Introducing automatically generated noisy descriptions would not alter this observation, but systematically evaluating their impact falls outside the scope of this work.

\textit{Conclusion:} Automatically rewritten prompts, even when designed to be more informative, rarely surpass the effectiveness of original human-curated narrative descriptions.

\textbf{RQ2.1: Among the strategies for prompt improvement, which yields the highest performance in narrative detection tasks?}

As in RQ1, we compute performance deltas between each generated prompt strategy and the Title baseline. These differences are then averaged across all model–dataset–task configurations to identify which strategies offer the most consistent improvements. For example, for the Self\_Gen\_Title strategy, we calculate the average of 32 F1-score differences, corresponding to 4 models $\times$ 8 evaluation points (2 languages in Dipromats and 3 classification schemes $\times$ 2 tasks in SemEval).

When comparing performance differences relative to Title-only prompts, \\
Self\_Gen\_Title\_Desc and GPT\_Gen\_Title\_Desc are the best-performing generated strategies, with average deltas of 1.89 pp and 1.87 pp, respectively. These approaches leverage both the title and the original description to construct a richer, yet controlled, narrative context.

This performance gain is intuitive: the improved prompts preserve the editorial framing of the original while refining it. Conversely, the Title\_Desc\_Guide variants—though semantically deeper—may introduce extraneous information that dilutes the core narrative signal.

\textit{Conclusion:} The most effective prompt improvements are those that augment, rather than radically reformulate, the original narrative context.

\textbf{RQ3: Does the narrative detection performance improve when using self-generated narrative descriptions, or is it preferable to rely on descriptions produced by more capable models, such as GPT systems?}

To address this research question, we directly compare prompt variants that differ only in the model used to generate the narrative descriptions, either self-generated by the classifier model itself or produced by a GPT-based system. Specifically, we calculate the F1-score differences between each pair of matching strategies, such as Self\_Gen\_Title vs.~GPT\_Gen\_Title, across all model and task configurations. These deltas are then averaged to assess whether one generation source consistently outperforms the other.

In Dipromats, no consistent performance advantage is observed between self-generated and GPT-generated descriptions across the Gen\_Title and Gen\_Title\_Desc strategies. However, for Gen\_Title\_Desc\_Guide, performance is similar across models, except in the case of Gemma, where self-generated prompts outperform GPT ones by 18 pp.

In SemEval, most differences are negligible. Two exceptions stand out: in the three-step classification setting, Self\_Gen\_Title exceeds GPT\_Gen\_Title by 5 pp, and Self\_Gen\_Title\_Desc\_Guide surpasses GPT\_Gen\_Title\_Desc\_Guide by 2.6 pp.

\textit{Conclusion:} On average, there is no significant difference, although there may be slight benefits to using self-generated descriptions.

\textbf{RQ4: How effective are ensemble strategies in improving the robustness and accuracy of zero-shot narrative detection?}

Across individual model ensembles (i.e., same model–prompt combinations across 5 runs), SemEval shows an average improvement of 2.9 pp. In Dipromats, majority-based ensembles offer a marginal 0.6 pp gain, while inclusion-based ensembles perform 0.6 pp worse.

In the cross-model ensemble setting, we evaluated combinations using all models (All) and excluding Granite (No\_Granite). Only the Title and OrigDesc prompt strategies were considered, given their superior performance and comparable results. To evaluate the effectiveness of ensemble strategies, we compared the average performance of model specific ensembles, created by aggregating five runs per model, with a cross-model ensemble combining predictions from all runs across models (15 for No\_Granite, 20 for All). 

In Dipromats, majority based ensembles consistently outperform the mean of model specific ensembles, often exceeding even the best individual run. In contrast, inclusion based ensembles underperform in all configurations, likely due to an increased rate of false positives when any of the 15 or 20 runs erroneously predicts an extra label.

In SemEval, majority based ensembles outperform the baseline in 62.5\% of cases, while inclusion ensembles improve results in only 20.8\%. Notably, inclusion ensembles degrade performance by up to 3.5 pp, particularly in subnarrative detection. Majority voting proves beneficial in this task, boosting performance by 8.5 pp with four models and 4.8 pp with three. However, even the best ensemble occasionally underperforms relative to top scoring individual runs, possibly due to the large output space and inherent difficulty of the task (evidenced by low interannotator agreement).

\textit{Conclusion:} Ensemble methods—particularly those based on majority voting substantially enhance robustness and accuracy, especially for fine-grained subnarrative detection, and can rival or surpass State Of The Art models trained with supervision.

\textbf{RQ5: How does the difference in model size affect performance in narrative identification?}

To assess model stability across prompt variations, we used the performance deltas computed in RQ1, i.e., the macro F1 differences between each prompt strategy and the Title strategy. By averaging these deltas separately for each classification model, we were able to estimate how sensitive each model is to prompt changes. This allows us to quantify performance volatility in relation to model size and architecture.

The Calme 78B model demonstrates the smallest performance variance across prompt strategies, with an average degradation of under 1 pp. In contrast, smaller models such as Gemma, Exaone, and Granite exhibit greater instability, with an average drop of 3.7 pp. In terms of ensemble stability, Gemma 12B is the most consistent across both datasets, followed by Calme 78B. Exaone and Granite show the greatest fluctuations.

Across all experiments, the standard deviation analysis reveals clear stability patterns among models and prompting strategies. Gemma and Calme consistently demonstrate the lowest variability, indicating robust and stable performance across tasks, while Exaone and especially Granite exhibit greater sensitivity to prompt structure and language context.

\textit{Conclusion:} Larger models are generally more robust to prompt variation, but some mid-sized models strike an effective balance between stability and computational cost.

\textbf{RQ6: Are the narrative descriptions generated by LLMs sufficiently coherent, semantically accurate, and pragmatically acceptable to serve as reliable inputs for narrative detection?}

A qualitative inspection suggests that most generated descriptions are semantically plausible and syntactically well-formed. However, prompts relying solely on titles may yield descriptions that deviate subtly from the annotators' intended framing, potentially altering narrative polarity or scope. 

\begin{tcolorbox}[title=Example of Semantic Drift in Generated Descriptions, colback=gray!5!white, colframe=gray!50!black, sharp corners=south]
\textbf{Generated Description EU6 (Granite, Using only Title):} \\
\textit{This narrative explores the European Union's (EU) positive impact on global affairs throughout 2020, focusing on its contributions to social, political, and cultural advancements. The EU navigates the challenges presented by the COVID-19 pandemic while maintaining its commitment to solidarity and cooperation, showcasing its role as a significant global player. The narrative likely delves into EU initiatives aimed at promoting healthcare equity, fostering economic resilience, safeguarding human rights, and preserving cultural heritage amidst the turmoil. Additionally, it may address the EU's efforts to mediate international conflicts and support sustainable development, emphasizing its multilateral approach to addressing global issues. This account underscores the EU's influence as a collective entity in shaping a more harmonious and equitable world, offering insights into its diverse member states' collaboration and unity during a tumultuous year.}

\vspace{0.3em}
\textbf{Original Human-Authored Descriptions:}
\begin{itemize}
    \item \textbf{EU3:} Classify tweets that promote EU policies and their benefits, demonstrating how European measures tangibly improve the lives of its citizens.
    \item \textbf{EU4:} Detect tweets that position the EU as a leader in tackling contemporary and future challenges such as climate change, digitalisation, and regulation.
    \item \textbf{EU6:} Find tweets that illustrate the EU’s contributions to global development, peace, safety, and well-being, emphasizing its role as a key partner for neighbouring and global societies.
\end{itemize}
\end{tcolorbox}

As shown, the generated text tends to blend multiple themes from related narratives and expand them beyond the scope of the original EU6 title. While this output remains plausible and linguistically fluent, it risks obscuring the intended boundaries of the target narrative, especially in multi-label or fine-grained classification settings.

\textit{Conclusion:} LLM-generated narrative descriptions can serve as reliable inputs, but their utility depends on alignment with the annotation framework. Overly abstract or generalised generations may diverge from ground-truth labelling logic.

\subsection{Confidence Intervals}
\begingroup
Quantifying confidence intervals is essential for validating claims regarding the robustness of the models. Performing formal statistical tests or deriving confidence intervals for aggregated scores would typically require a large number of independent runs, which is computationally impractical. An alternative approach, inspired by \cite{fraile-hernandezMeasuringLargeLanguage2025}, allows for approximate confidence interval estimation using a resampling methodology. This method leverages the gold-standard labels together with the model predictions for each instance to perform bootstrap sampling with replacement, calculating a score for each resampled set. The resulting distribution of scores is then used to determine the 95\% confidence interval via the percentile method.

As the gold-standard labels for SemEval are not publicly available, this analysis has been conducted on the Dipromats dataset. For each model and prompting strategy, the predictions from the five independent runs were concatenated, and 2000 bootstrap samples of 800 instances (equal to the test set size) were generated. The resulting 95\% confidence intervals for each configuration are presented in Table~\ref{tab:confidence_intervals}.

\begin{table}[h]
\centering
\footnotesize
\begin{tabular}{lccccc} 
\hline
      & \textbf{Lang} & \textbf{Calme}                   & \textbf{Gemma}                   & \textbf{Exaone}                  & \textbf{Granite}                  \\[1.25ex]
\hline
\multirow{2}{*}{\textbf{Title}}                                                                          & es            & (0.612, 0.657)                   & (0.591, 0.634)                   & (0.528, 0.575)                   & (0.306, 0.359)                    \\
      & en            & (0.560, 0.606)                   & (0.544, 0.585)                   & (0.474, 0.522)                   & \underline{(0.345, 0.391)}  \\[1.25ex]
\hline
\multirow{2}{*}{\textbf{OrigDesc}}                                                                       & es            & (0.615, 0.662)                   & \underline{(0.601, 0.648)} & \underline{(0.548, 0.597)} & \underline{(0.310, 0.359)}  \\
      & en            & (0.591, 0.637)                   & \underline{(0.572, 0.613)} & \underline{(0.512, 0.557)} & (0.328, 0.375)                    \\[1.25ex]
\hline
\multirow{2}{*}{\textbf{Self\_Gen\_Title}}                                                               & es            & (0.599, 0.643)                   & (0.555, 0.598)                   & (0.483, 0.530)                   & (0.264, 0.309)                    \\
      & en            & (0.548, 0.593)                   & (0.511, 0.552)                   & (0.462, 0.505)                   & (0.276, 0.320)                    \\[1.25ex] 
\hline
\multirow{2}{*}{\begin{tabular}[c]{@{}l@{}}\textbf{Self\_Gen\_}\\\textbf{Title\_Desc}\end{tabular}}     & es            & (0.608, 0.656)                   & (0.564, 0.608)                   & (0.522, 0.569)                   & (0.266, 0.311)                    \\
      & en            & \underline{(0.596, 0.641)} & (0.549, 0.591)                   & (0.488, 0.537)                   & (0.306, 0.352)                    \\[1.25ex] 
\hline
\multirow{2}{*}{\begin{tabular}[c]{@{}l@{}}\textbf{Self\_Gen\_Title}\\\textbf{Desc\_Guide}\end{tabular}} & es            & (0.607, 0.656)                   & (0.512, 0.557)                   & (0.432, 0.483)                   & (0.221, 0.270)                    \\
      & en            & (0.583, 0.629)                   & (0.455, 0.500)                   & (0.402, 0.451)                   & (0.260, 0.306)                    \\[1.25ex] 
\hline
\multirow{2}{*}{\textbf{GPT\_Gen\_Title}}                                                                & es            & (0.586, 0.632)                   & (0.563, 0.606)                   & (0.502, 0.548)                   & (0.245, 0.290)                    \\
      & en            & (0.542, 0.589)                   & (0.502, 0.545)                   & (0.470, 0.512)                   & (0.275, 0.320)                    \\[1.25ex] 
\hline
\multirow{2}{*}{\begin{tabular}[c]{@{}l@{}}\textbf{GPT\_Gen\_}\\\textbf{Title\_Desc}\end{tabular}}      & es            & \underline{(0.618, 0.663)} & (0.592, 0.637)                   & (0.524, 0.568)                   & (0.261, 0.309)                    \\
      & en            & (0.592, 0.637)                   & (0.559, 0.599)                   & (0.503, 0.546)                   & (0.276, 0.324)                    \\[1.25ex] 
\hline
\multirow{2}{*}{\begin{tabular}[c]{@{}l@{}}\textbf{GPT\_Gen\_Title}\\\textbf{Desc\_Guide}\end{tabular}}  & es            & (0.603, 0.651)                   & (0.240, 0.310)                   & (0.460, 0.509)                   & (0.198, 0.243)                    \\
      & en            & (0.582, 0.627)                   & (0.347, 0.412)                   & (0.414, 0.461)                   & (0.249, 0.294)                   
\end{tabular}
\caption{95\% confidence intervals for all model and prompting configurations on the Dipromats test set}
\label{tab:confidence_intervals}
\end{table}

Across models, languages, and prompting strategies, the 95\% confidence intervals exhibit broadly comparable widths, indicating a similar degree of variability in performance. To identify the strongest configurations, it is therefore most appropriate to focus on the interval midpoints. Notably, the configurations with the highest midpoints correspond exactly to those that achieve the best aggregated mean F1 scores in Table~\ref{tab:dipromats_results}, confirming the stability of the rankings obtained from averaging the five runs. This alignment suggests that the aggregated means provide a reliable basis for comparative claims.

\endgroup

\subsection{Qualitative Error Analysis}

\label{sec:error_analysis}
To further interpret the quantitative findings, a qualitative error analysis was conducted to examine specific cases where the models succeed or fail in identifying narratives. This analysis is only feasible for the \textit{Dipromats} dataset, as it is the only collection for which gold-standard labels and the original English tweets are available. Only tweets containing at least one \textit{yes} label for any narrative were considered. The goal is to explore the types of errors that arise under different prompting strategies and model configurations, as well as to understand how narrative descriptions influence classification outcomes.

\subsubsection{Cross-model and cross-prompt variability}
The analysis jointly considers predictions from the four evaluated models, each tested with eight prompting strategies and five independent runs, yielding a total of 160 predictions per tweet.  
Tweets such as the one shown in Figure~\ref{box:lowagreement} were correctly classified in only one out of 160 predictions. This illustrates the high degree of subjectivity in some tweets, where the stance toward a narrative is subtle, implicit, or context-dependent. Overall, 26 out of the 562 instances were correctly identified in fewer than 10\% of all predictions, reflecting intrinsic ambiguity.

\begin{figure}
\footnotesize
    \centering
\begin{tcolorbox}[]
\textbf{Tweet:}\\
\textit{This morning we worked on the Commission's response to \#coronavirus in view of the EUCO teleconference this afternoon. Also called @GiuseppeConteIT this morning to express my support for the Italian people. We'll 'video-meet' tomorrow; discuss national measures and EU support.}\\
\textbf{Supported Narrative:}
\begin{itemize}
    \item \textbf{EU4:} The European Union is an avant-garde political actor. Detect tweets that position the EU as a leader in tackling contemporary and future challenges such as climate change, digitalisation, and regulation.
\end{itemize}

\end{tcolorbox}
    \caption{A tweet for which only 1 out of 160 predictions across all models and prompt strategies is correct.}
\label{box:lowagreement}
\end{figure}

In contrast, tweets such as the one illustrated in Figure~\ref{box:calmeonly} achieved 100\% accuracy across all runs performed with the Calme model, yet displayed markedly lower performance for the remaining models.  
In the opposite case, tweets like that in Figure~\ref{box:smallmodelsucceed} show the reverse pattern: the largest model fails across all runs, while smaller models achieve correct predictions. However, even in these cases, the accuracy for the smaller models rarely exceeds 30\%.
\begin{figure}
\footnotesize
    \centering
\begin{tcolorbox}[]
\textbf{Tweet:}\\
\textit{Under U.S. law and practice those who participate and benefit from electoral fraud, undermine democratic institutions and impede a peaceful transition of power can be subject to a variety of consequences. \#Guyana}\\
\textbf{Supported Narrative:}
\begin{itemize}
    \item \textbf{US4:} The US is a force for good. Showcases American support for human rights, freedoms, oppressed peoples, and its efforts toward peace, development, and security.
\end{itemize}

\end{tcolorbox}
    \caption{A tweet correctly classified in all runs by the largest model (Calme) but showing substantially lower accuracy for smaller models.}
\label{box:calmeonly}
\end{figure}

\begin{figure}
\footnotesize
    \centering
\begin{tcolorbox}[]
\textbf{Tweet:}\\
\textit{President Vladimir \#Putin  @UNGA75: \#Russia is completely open to partner relations and willing to cooperate. We are proposing to hold an online High-Level Conference shortly for countries interested in cooperation in the development of anti \#COVID19 vaccine.}\\
\textbf{Supported Narrative:}
\begin{itemize}
    \item \textbf{RU2:} Russia leads an alternative system to that sponsored by the West. Showcases Russia's advocacy for a fairer, more multilateral system, promoting cooperation outside Western influence. Non-Western cooperation is highlighted as an example Russia upholds true multilateralism, based on the respect of national sovereignty.
\end{itemize}

\end{tcolorbox}
    \caption{A tweet that the largest model fails to classify correctly, whereas smaller models achieve partial success.}
\label{box:smallmodelsucceed}
\end{figure}
\subsubsection{Focusing on the Calme model: the role of narrative descriptions.}
A deeper inspection was carried out for the Calme model to evaluate the influence of narrative prompting. Specifically, performance differences between using only the narrative title (Title) and other prompting strategies involving extended descriptions (original or automatically generated) were examined.

Two representative examples are presented in Figure~\ref{box:titlewins}, which are the only tweets in the dataset for which all five runs using only the Title prompt correctly identify the narratives, while none of the remaining 35 runs (those employing full or generated descriptions) produce the correct label.
\begin{figure}
\footnotesize
    \centering
\begin{tcolorbox}[]
\textbf{Tweets:}\\
1. \sep \textit{\#Lavrov: There is a joke that the \#US will never have a colour revolution because it doesn’t have an \#American Embassy. Every joke has some truth in it. However, we would not want a leading global power like the US to fall into a deep crisis.}\\
2. \sep \textit{The U.S. Embassy joins our fellow Americans and people around the world in mourning the loss of U.S. Supreme Court Justice Ruth Bader Ginsburg. She was an inspiration not only to countless women and girls, but everyone who looked up to our Constitution and rule of law.}\\
\textbf{Supported Narrative:}
\begin{itemize}
    \item[1.] \textbf{RU3:} Russia is a benevolent partner. Emphasizes Russia's role in promoting peace, solidarity, and development, positioning it as a reliable partner for just causes.
    \item[2.] \textbf{US2:} The US has an admirable society. Promotes the different ethnic groups of the country, its civil society, its companies and its military heroes.
\end{itemize}
\end{tcolorbox}
    \caption{Two tweets correctly classified in all Title-only runs but misclassified in all other runs using original or generated descriptions.}
\label{box:titlewins}
\end{figure}
Analogously, Figure~\ref{box:descwins} displays the only two tweets that exhibit the opposite behaviour—all 35 runs using narrative descriptions achieve correct predictions, whereas none of the five Title-only runs succeed.\begin{figure}
\footnotesize
    \centering
\begin{tcolorbox}[]
\textbf{Tweets:}\\
1. \sep \textit{We have an ambitious target to be climate neutral by 2050. EVP @TimmermansEU and U.S. @ClimateEnvoy John Kerry met to discuss our goals for @COP26. Follow their press statements \#EUGreenDeal}\\
2. \sep \textit{President Xi Jinping: Economic globalization; win-win cooperation is still the trend of the times. \#China will continue to deepen reform and opening-up, strengthen cooperation, promote an open world economy, and push for building a community with a shared future for mankind.}\\
\textbf{Supported Narrative:}
\begin{itemize}
    \item[1.] \textbf{EU4:} The European Union is an avant-garde political actor. Detect tweets that position the EU as a leader in tackling contemporary and future challenges such as climate change, digitalisation, and regulation.
    \item[2.] \textbf{CH2:} China is a benevolent power. This narrative highlights China's cooperative stance, emphasizing support for justice, international law, security, multilateralism and economic development in other nations, and pursuit of mutual benefits. Through this narrative, China may promote a system in which every country will participate and contribute to decide the fate of the world. One of the Chinese motos is a community of a shared destiny for mankind.
\end{itemize}
\end{tcolorbox}
    \caption{Two tweets correctly classified in all runs using original or generated descriptions but misclassified in all Title-only runs.}
\label{box:descwins}
\end{figure}

To summarise these contrasts numerically, Table~\ref{tab:title_vs_desc_conditions} reports the number of instances meeting each of the following conditions:

\begin{itemize}
    \item \textbf{C1:} All Title runs are correct and all compared-prompt runs fail.
    \item \textbf{C2:} All Title runs fail and all compared-prompt runs are correct.
    \item \textbf{C3:} Some Title runs succeed while all compared-prompt runs fail.
    \item \textbf{C4:} All Title runs fail while some compared-prompt runs succeed.
\end{itemize}
\begin{table}[]
\centering
\begin{tabular}{l|cccc}
                              & \textbf{C1} & \textbf{C2} & \textbf{C3} & \textbf{C4} \\ \hline
\textbf{OrigDesc}                      & 11 & 20 & 39 & 35 \\
\textbf{Self\_Gen\_Title}              & 15 & 12 & 46 & 30 \\
\textbf{Self\_Gen\_Title\_Desc}        & 20 & 27 & 43 & 44 \\
\textbf{Self\_Gen\_Title\_Desc\_Guide} & 28 & 22 & 60 & 37 \\
\textbf{GPT\_Gen\_Title}               & 16 & 14 & 50 & 33 \\
\textbf{GPT\_Gen\_Title\_Desc}         & 13 & 22 & 39 & 37 \\
\textbf{GPT\_Gen\_Title\_Desc\_Guide}  & 27 & 33 & 56 & 49
\end{tabular}
\caption{Comparison between Title-only and other prompting strategies for the Calme model.}
\label{tab:title_vs_desc_conditions}
\end{table}
The numerical contrasts in Table~\ref{tab:title_vs_desc_conditions} indicate that incorporating narrative descriptions—whether original or automatically generated—can substantially alter the predictive behaviour compared to using the title alone. For instance, richer prompt strategies such as Self\_Gen\_Title\_Desc\_Guide and GPT\_Gen\_Title\_Desc\_Guide exhibit higher counts in C2 and C4, reflecting cases where these descriptions enable correct predictions that the title-only runs fail to capture. Conversely, elevated values in C1 suggest that in some instances the additional descriptions may actually reduce performance relative to the concise title, potentially due to noise or over-specification introduced by the model-generated content.

\section{Comparison with the State of the Art}
\label{sec:comparison_sota}
To provide a head-to-head comparison with state-of-the-art systems, Tables~\ref{tab:semeval_headtohead} and~\ref{tab:dipromats_headtohead} report the performance of our best zero-shot model alongside the top leaderboard submissions from SemEval and Dipromats, respectively. In both tables, $\Delta$ denotes the difference in performance relative to the best-performing system.

\begin{table}[h]
\centering

\begin{tabular}{l|c|cc|cc}
\textbf{System} & \textbf{Setting} & \textbf{F1 nar} & \textbf{F1 subnar} & $\Delta$ \textbf{F1 nar} & $\Delta$ \textbf{F1 subnar} \\ \hline
\underline{GATENLP}      & \underline{Supervised} & \underline{0.6270} & \underline{0.4630} & --    & --    \\
COGNAC       & Zero-Shot  & 0.5540 & 0.4260 & -0.073 & -0.037 \\
INSALyon2    & Zero-Shot  & 0.5130 & 0.4060 & -0.114 & -0.057 \\
\textit{Our System} & \textit{Zero-Shot} & \textit{0.5090} & \textit{0.4000} & \textit{-0.118} & \textit{-0.063} \\
iLostTheCode & Supervised & 0.4980 & 0.3730 & -0.129 & -0.090 \\
KostasThesis & Supervised & 0.5560 & 0.3620 & -0.071 & -0.101 \\
NCLteam      & Supervised & 0.4860 & 0.3450 & -0.141 & -0.118 \\
Narrlangen   & Zero-Shot  & 0.4440 & 0.3440 & -0.183 & -0.119 \\
PATeam       & Supervised & 0.5210 & 0.3390 & -0.106 & -0.124 \\
PingAnAI     & Supervised & 0.5210 & 0.3390 & -0.106 & -0.124 \\
\end{tabular}
\caption{Head-to-head comparison of our best zero-shot system against leaderboard systems on SemEval 2025 Task 10 (Subtask 2).}
\label{tab:semeval_headtohead}
\end{table}

\begin{table}[h]
\centering
\begin{tabular}{l|c|cc|cc}
\textbf{System} & \textbf{Setting} & \textbf{F1 es} & \textbf{F1 en} & $\Delta$ \textbf{F1 es} & $\Delta$ \textbf{F1 en} \\ \hline
\underline{OB1}         & \underline{Few-Shot} & \underline{0.6854} & \underline{0.6701} & --     & --     \\
UNED-JF     & Few-Shot & 0.6847 & 0.6367 & -0.0007 & -0.0334 \\
\textit{Our System} & \textit{Zero-Shot} & \textit{0.6453} & \textit{0.6209} & \textit{-0.0401} & \textit{-0.0492} \\
ManchaAzul  & Few-Shot Multi-Agent   & 0.6403 & 0.6117 & -0.0451 & -0.0584 \\
UNED-AP     & Linear Transformation & 0.4993 & 0.5226 & -0.1861 & -0.1475 \\
UMU-Team    & Few-Shot   & 0.3736 & 0.4207 & -0.3118 & -0.2494 \\
Baseline    & Zero-Shot  & 0.3769 & 0.4160 & -0.3085 & -0.2541 \\
\end{tabular}
\caption{Head-to-head comparison of our best zero-shot system against leaderboard systems on Dipromats 2024 Task 2.}
\label{tab:dipromats_headtohead}
\end{table}

For SemEval, participating systems covered both supervised and zero-shot strategies. The GATENLP team \cite{singhGateNLPSemEval2025Task2025} achieved the highest overall performance by implementing a hierarchical three-step prompting framework combined with fine-tuning of a \texttt{LLaMa 3.2} model \cite{grattafioriLlama3Herd2024}. Their results demonstrate the clear advantage of utilising task-specific training data, as their zero-shot experiments yielded substantially lower scores. The COGNAC \cite{grattafioriLlama3Herd2024} system adopted a different strategy, first summarising news articles into texts of 300 words or fewer, while explicitly retaining key topics, sentiments, and narratives. Narrative detection was then formulated as a binary classification task, followed by subnarrative detection for those narratives identified as supported. Both stages were implemented in a zero-shot setting using \texttt{GPT-4o-mini} and \texttt{LLaMa 3.1-8B-Instruct}. However, their results are not fully comparable to other submissions, as no explicit topic classification module was reported; it can be inferred that topic identification was carried out using the original filenames of the news files. In our own experiments, topic classification accuracy on the training and development sets ranged between 0.8 and 0.9, highlighting the potential impact of this assumption on overall performance. The INSALyon2 \cite{eljadiriTeamINSALyon2SemEval2025} team also adopted a zero-shot methodology, employing a multi-agent framework based on \texttt{GPT-4o} and \texttt{GPT-4o-mini}. By contrast, the remaining participants relied on supervised approaches, training their systems with the task-specific data provided. Within this context, our proposed system demonstrates competitive performance without the need for any training data, reinforcing its effectiveness as a zero-shot solution.

In the Dipromats leaderboard, the OB1 team reports the highest score, although no details are provided regarding the underlying model or methodological choices, which limits comparability. The UNED-JF team employed the \texttt{calme-2.4-rys-78b} model quantized to 8 bits and adopted a few-shot strategy using only the titles of the narratives. The ManchaAzul team \cite{caballeroLLMBasedMultiAgentModels} adopted a multi-agent few-shot strategy with \texttt{GPT-4}, while UNED-AP transformed the vector space of tweet embeddings to perform classification. The official baseline relied on a simple zero-shot approach using \texttt{Mixtral 8x7B} \cite{jiangMixtralExperts2024}, and the UMU-Team \cite{garcia-diazUMUTEAMDIPROMATS20242024} implemented a few-shot strategy with \texttt{Zephyr-7B-beta} \cite{tunstallZephyrDirectDistillation2023}. Within this competitive landscape, our system achieves robust performance, surpassing models based on \texttt{GPT-4} despite its comparatively lower computational requirements.

\section{Generalisation of Results to Other Languages}
\label{sec:generalisation_languages}
\begingroup
Although our main experiments are conducted in a limited set of languages, it is important to assess whether the proposed unsupervised narrative detection framework generalises beyond this setting. Leveraging the multilingual capabilities of large generative models, we examine whether our previous conclusions hold when analysing texts in other languages. To this end, we consider two complementary settings: providing the models with documents in their original language, and supplying translated versions of the same texts as input.

\subsection{Experimental setting}
To investigate the generalisation of our approach to other languages, we use the development set of the SemEval dataset, which includes documents written in Bulgarian, Portuguese, Hindi, and Russian. The distribution of instances per language is as follows: 35 documents each in Bulgarian, Portuguese, and Hindi, and 32 in Russian.

For this exploratory evaluation, we employ the Calme 78B model and the hybrid classification approach, as it achieved the strongest performance in the experiments presented in Section~\ref{sec:results}. We explore two alternative input strategies. In the first setting, the model is provided with the news article in its original language, allowing us to evaluate whether it can directly leverage its multilingual representations to identify narratives without any language normalisation. In the second setting, the same articles are automatically translated into English using OPUS models~\cite{tiedemannOPUSMTBuildingOpen2020} and then used as input to the model. This second approach allows us to study transfer to languages that the model has not been exposed to during training. 

Given that the aim of this analysis is to provide a lightweight yet informative multilingual slice rather than a full-scale replication of the pipeline, we run a single inference pass per document instead of the five-run setting used elsewhere in the paper.

\subsection{Feeding models with input texts in other languages}

Table~\ref{tab:crosslingual_results_original} summarises the model’s performance when using the documents in their original languages (Bulgarian, Portuguese, Hindi, and Russian). For each configuration, the first column reports the F1 score for narrative prediction, while the second column reports the F1 score for subnarrative detection.

\begin{table}[h]
\centering
\footnotesize
\begin{tabular}{l|rr|rr|rr|rr}
& \multicolumn{2}{c|}{\textbf{BG}}                                              & \multicolumn{2}{c|}{\textbf{PT}}                                              & \multicolumn{2}{c|}{\textbf{HI}}                                              & \multicolumn{2}{c}{\textbf{RU}}                                              \\[1.25ex]
\textbf{Title}                                                                    & 0.5276                              & \underline{0.3659}                         & 0.4314                              & 0.2057                                  & 0.4007                              & 0.2244                                  & 0.5375                              & 0.3382                                 \\[1.25ex] \hline
\textbf{OrigDesc}                                                                 & 0.5124                              & 0.3186                                  & \underline{0.5295}                     & \underline{0.2129}                         & 0.4101                              & \underline{0.2528}                         & \underline{0.5983}                     & \underline{0.3383}                        \\[1.25ex] \hline
\textbf{Self\_Gen\_Title}                                                         & 0.5571                              & 0.3477                                  & 0.3805                              & 0.1129                                  & 0.3720                              & 0.1940                                  & 0.5012                              & 0.2713                                 \\[1.25ex] \hline
\textbf{\begin{tabular}[c]{@{}l@{}}Self\_Gen\_\\ Title\_Desc\end{tabular}}        & \underline{0.6076}                     & 0.3524                                  & 0.3948                              & 0.1333                                  & \underline{0.4222}                     & 0.1962                                  & 0.5321                              & 0.2494                                 \\[1.25ex] \hline
\textbf{\begin{tabular}[c]{@{}l@{}}Self\_Gen\_\\ Title\_Desc\_Guide\end{tabular}} & 0.5629                              & 0.3015                                  & 0.3981                              & 0.1610                                  & 0.3981                              & 0.2467                                  & 0.5316                              & 0.2657                                 \\[1.25ex] \hline
\textbf{GPT\_Gen\_Title}                                                          & 0.5000                              & 0.2877                                  & 0.4133                              & 0.1571                                  & 0.3773                              & 0.1994                                  & 0.5114                              & 0.2540                                 \\[1.25ex] \hline
\textbf{\begin{tabular}[c]{@{}l@{}}GPT\_Gen\_\\ Title\_Desc\end{tabular}}         & 0.5600                              & 0.3467                                  & 0.4097                              & 0.1034                                  & 0.3392                              & 0.2112                                  & 0.5005                              & 0.2735                                 \\[1.25ex] \hline
\textbf{\begin{tabular}[c]{@{}l@{}}GPT\_Gen\_\\ Title\_Desc\_Guide\end{tabular}}  & 0.4924                              & 0.2273                                  & 0.3649                              & 0.1533                                  & 0.3554                              & 0.1673                                  & 0.5158                              & 0.3113                                
\end{tabular}
\caption{Evaluation results using the original language documents on the SemEval development set.}
\label{tab:crosslingual_results_original}
\end{table}

\subsection{Feeding models with translations into English}
Table~\ref{tab:crosslingual_results_en} presents the results when the documents are translated into English. The same reporting scheme applies: the first column indicates the F1 score for narrative detection, and the second column the F1 score for subnarrative identification.
\begin{table}[h]
\footnotesize
\centering
\begin{tabular}{l|cc|cc|cc|cc}
& \multicolumn{2}{c|}{\textbf{BG}}                    & \multicolumn{2}{c|}{\textbf{PT}}                    & \multicolumn{2}{c|}{\textbf{HI}}                    & \multicolumn{2}{c}{\textbf{RU}}                      \\[1.25ex] 
\textbf{Title}                                                                             & \underline{0.6078} & \underline{0.4248} & 0.4771                   & \underline{0.2257} & 0.3329                   & 0.2305                   & \underline{0.6691} & \underline{0.4312}  \\[1.25ex]  
\hline
\textbf{OrigDesc}                                                                          & 0.5695                   & 0.3356                   & 0.4062                   & 0.1635                   & \underline{0.4016} & \underline{0.2939} & 0.4432                   & 0.3057                    \\[1.25ex] 
\hline
\textbf{Self\_Gen\_Title}                                                                  & 0.5659                   & 0.3729                   & 0.4124                   & 0.1659                   & 0.3473                   & 0.2080                   & 0.5050                   & 0.2712                    \\[1.25ex]  
\hline
\begin{tabular}[c]{@{}l@{}}\textbf{Self\_Gen\_}\\\textbf{Title\_Desc}\end{tabular}        & 0.5263                   & 0.2997                   & 0.3301                   & 0.1502                   & 0.3109                   & 0.1852                   & 0.5308                   & 0.2720                    \\[1.25ex]  
\hline
\begin{tabular}[c]{@{}l@{}}\textbf{Self\_Gen\_Title\_}\\\textbf{Desc\_Guide}\end{tabular} & 0.5524                   & 0.3003                   & 0.3962                   & 0.1834                   & 0.3724                   & 0.2060                   & 0.5222                   & 0.2937                    \\[1.25ex]  
\hline
\textbf{GPT\_Gen\_Title}                                                                   & 0.5452                   & 0.3702                   & 0.4350                   & 0.1610                   & 0.3074                   & 0.1727                   & 0.4915                   & 0.2399                    \\[1.25ex]  
\hline
\begin{tabular}[c]{@{}l@{}}\textbf{GPT\_Gen\_}\\\textbf{Title\_Desc}\end{tabular}         & 0.5563                   & 0.3710                   & \underline{0.4873} & 0.1529                   & 0.3750                   & 0.2166                   & 0.5292                   & 0.2135                    \\[1.25ex]  
\hline
\begin{tabular}[c]{@{}l@{}}\textbf{GPT\_Gen\_Title\_}\\\textbf{Desc\_Guide}\end{tabular}  & 0.5743                   & 0.3210                   & 0.3135                   & 0.1105                   & 0.2718                   & 0.1532                   & 0.4820                   & 0.2334                   
\end{tabular}
\caption{Evaluation results using English translations of the SemEval development set documents.}
\label{tab:crosslingual_results_en}
\end{table}
\subsection{Results}
Across both original and translated documents, a clear pattern emerges: the most effective prompting strategies are consistently those that use only the narrative title or combine the title with the original human-authored description. Automatically generated descriptions do not appear to offer a meaningful improvement over these simpler strategies. This suggests that the large language models possess sufficient contextual and pragmatic knowledge to identify narratives without relying on augmented or automatically generated narrative descriptions.

These findings reinforce the earlier conclusions drawn from the English and Spanish experiments, indicating that human-curated descriptions provide a robust signal, whereas additional automatically generated content does not systematically enhance zero-shot classification performance.

\endgroup
\section{Scalability Analysis}
\label{sec:scalability}
\begingroup
To assess the practical feasibility of the proposed zero-shot narrative detection methodology, a limited scalability analysis was conducted. While the main focus of this study is on evaluating the intrinsic capabilities of LLMs in narrative identification, it is important to characterise the computational requirements and efficiency of different models and prompt strategies.

Tables~\ref{tab:dipromats_scalability} and~\ref{tab:semeval_scalability} summarise the computational footprint for the OrigDesc prompt strategy across the four models. For each dataset, the number of requests sent to the model, the average time per request, and the total GPU time are reported. All experiments were executed on a high-end workstation with 4 $\times$ NVIDIA RTX A5000 GPUs (24\,GB GDDR6 each) and 256\,GB of system RAM.

\begin{table}[h]
\centering
\begin{tabular}{c|ccc}
\textbf{Model} & \textbf{\# Req} & \textbf{\begin{tabular}[c]{@{}c@{}}Avg Time/Req\\ (s)\end{tabular}} & \textbf{\begin{tabular}[c]{@{}c@{}}Total GPU\\ Time (s)\end{tabular}} \\ \hline
Calme          & 1600            & 18                                                                  & 28800                                                                 \\
Gemma          & 1600            & 8                                                                   & 12800                                                                 \\
Exaone         & 1600            & 5                                                                   & 8000                                                                  \\
Granite        & 1600            & 3                                                                   & 4800                                                                 
\end{tabular}
\caption{Summary of the OrigDesc processing statistics for the Dipromats dataset.}
\label{tab:dipromats_scalability}
\end{table}

\begin{table}[h]
\centering
\begin{tabular}{l|lccc}
\textbf{Model}       & \textbf{\begin{tabular}[c]{@{}c@{}}Classification\\ Scheme\end{tabular}} & \textbf{\# Req} & \textbf{\begin{tabular}[c]{@{}c@{}}Avg Time/Req\\ (s)\end{tabular}} & \multicolumn{1}{c}{\textbf{\begin{tabular}[c]{@{}c@{}}Total GPU\\ Time (s)\end{tabular}}} \\ \hline
                     & All-in-One                                                               & 101             & 61                                                                  & \multicolumn{1}{c}{6180}                                                                  \\
Calme                & Hybrid                                                                   & 202             & 52                                                                  & \multicolumn{1}{c}{10504}                                                                 \\
\multicolumn{1}{l|}{} & 3-Step                                                                   & 417             & 44                                                                  & \multicolumn{1}{c}{18348}                                                                 \\ \hline
                     & All-in-One                                                               & 101             & 23                                                                  & 2323    \\
Gemma                & Hybrid                                                                   & 202             & 17                                                                  & 3434    \\
\multicolumn{1}{l|}{} & 3-Step                                                                   & 543             & 14                                                                  & 7602    \\ \hline
                     & All-in-One                                                               & 101             & 18                                                                  & 1818    \\
Exaone               & Hybrid                                                                   & 202             & 15                                                                  & 3030    \\
\multicolumn{1}{l|}{} & 3-Step                                                                   & 401             & 9                                                                   & 3609    \\ \hline
                     & All-in-One                                                               & 101             & 7                                                                   & 707     \\
Granite              & Hybrid                                                                   & 202             & 5                                                                   & 1010    \\
\multicolumn{1}{l|}{} & 3-Step                                                                   & 482             & 3                                                                   & 1446   
\end{tabular}
\caption{Summary of the OrigDesc processing statistics for the SemEval dataset.}
\label{tab:semeval_scalability}
\end{table}

To illustrate the variability in input length, Figure~\ref{fig:semeval_tokens} shows the distribution of the number of tokens per document for the OrigDesc prompt strategy on SemEval using All-in-One approach, across all four models. While Dipromats tweets are short and exhibit limited variability in token counts, SemEval news articles display a wider range, which affects both memory usage and inference latency.

\begin{figure}[h]
    \centering
    \includegraphics[width=0.99\linewidth]{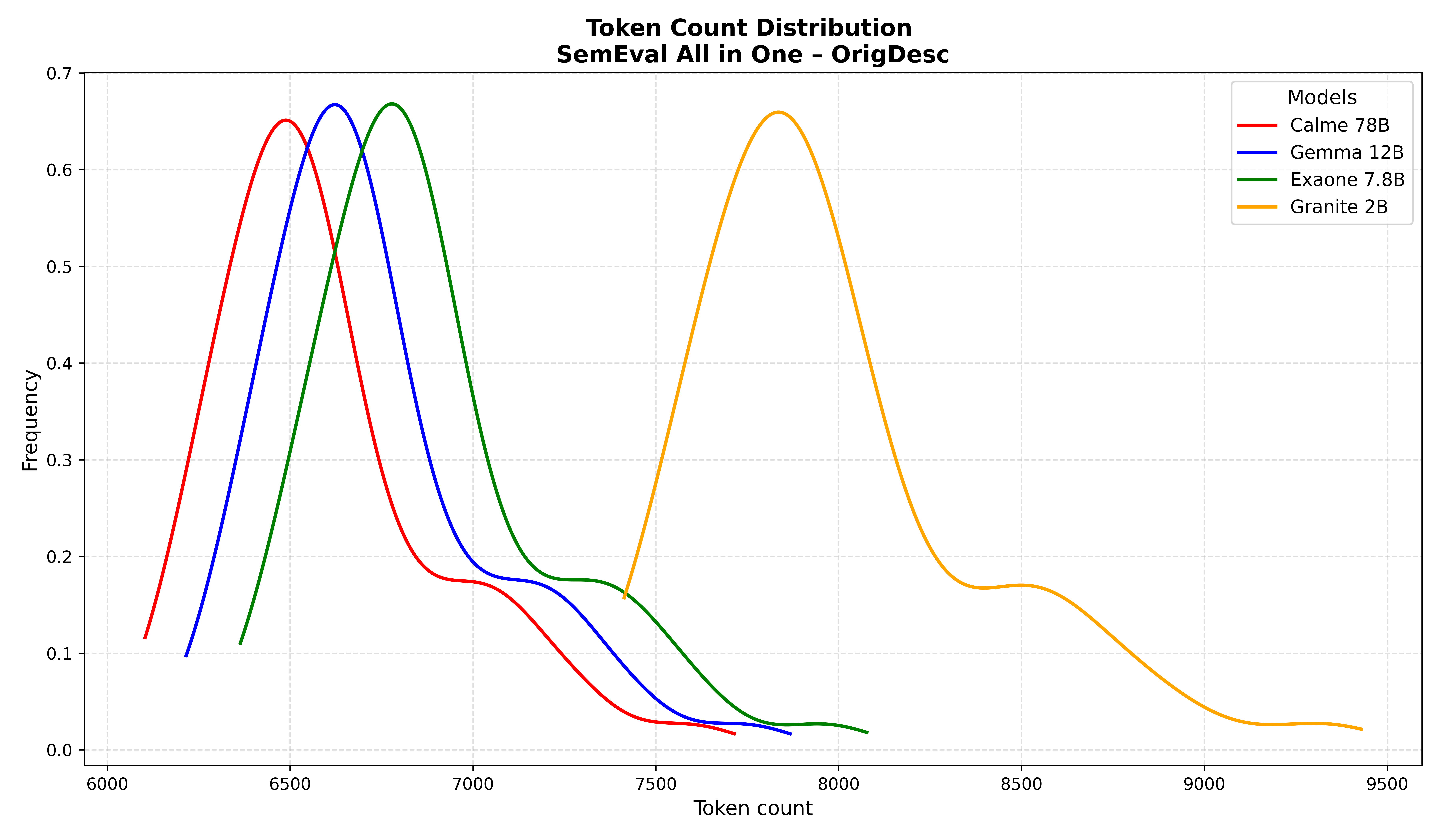}
    \caption{Distribution of the number of input tokens per document for the \textit{OrigDesc} prompt strategy on the \textit{SemEval} dataset across the four evaluated models.}

    \label{fig:semeval_tokens}
\end{figure}

As shown in Figure~\ref{fig:semeval_tokens}, it can be observed that, as model size decreases, the number of tokens that needs to be processed increases, reflecting differences in how models encode and interpret the narrative context.  

\endgroup

\section{Conclusion}
\label{sec:conclusion}

This study explores the ability of large pre-trained linguistic models to interpret and classify the underlying narratives in social messages without the need for task-specific training. Our zero‑shot experiments across the Dipromats and SemEval datasets confirm that the provision of human-written narrative descriptions, consistently enhances performance compared to title‑only prompts. While automatically generated descriptions are grammatically fluent and often semantically rich, they frequently introduce subtle shifts in framing that can degrade classification accuracy, especially in multilabel and fine‑grained subnarrative scenarios.

Ensemble strategies, and in particular majority‑voting across diverse model runs, further bolster robustness and frequently surpass the best individual model performances. The greatest gains appear in subnarrative detection, where aggregation reduces the impact of model-specific variability. Importantly, excluding the smallest model from the ensemble yields a more stable outcome, indicating a need to balance diversity with baseline reliability.

Model scale emerges as a key factor: the largest architectures not only exhibit the least sensitivity to prompt variation, but also achieve the highest overall performance, yet certain mid‑sized models (such as 12B parameters) achieve comparable stability at a lower computational cost. This suggests that, in resource‑constrained settings, carefully chosen mid‑scale models may offer an effective compromise.

Furthermore, a specific multilingual evaluation conducted with the SemEval development dataset shows that the narrative detection framework without prior training maintains promising performance in Bulgarian, Portuguese, Hindi, and Russian, indicating that the approach could be generalised beyond English and would support the broader applicability of the proposed methodology.

In conclusion, our findings validate the hypothesis that LLMs encode sufficient pragmatic and contextual knowledge to identify strategic narratives in zero‑shot settings. When guided by human-written coherent descriptions and combined through simple ensembling, these models rival supervised systems, offering a scalable alternative for narrative detection in domains without training data.

\section{Future Work}
\label{sec:future_work}
An interesting extension of the present study would be to expand the analysis to multiple languages and investigate cross-lingual transfer. While the current work focuses primarily on English tweets and news articles, evaluating LLMs’ zero-shot narrative detection performance across languages could provide additional insights into their generalisation capabilities.

One promising direction is the incorporation of few-shot learning setups. While the current work is strictly zero-shot, our results suggest that modest gains could be achieved through minimal supervision. Given that the cost of producing high-quality narrative annotations is relatively low compared to full dataset labelling, future research could explore the performance in few-shot scenarios, particularly in complex subnarrative detection tasks.

Another avenue involves the adoption of multi-agent architectures. Rather than relying on a single general-purpose model, future systems could benefit from ensembles of specialised agents, each prompted to focus on specific narrative features, such as stance, polarity, topical alignment, or temporal framing. These division of labour may facilitate more robust narrative identification, especially in multi-label and noisy environments.

Beyond these directions, future work could explore human-in-the-loop frameworks to support the generation of narrative descriptions. While automatic generation methods offer scalability, they may lack alignment with annotators’ interpretative frameworks or the dataset's underlying taxonomy. Introducing human supervision may improve semantic fidelity and task relevance.

In addition, a systematic analysis of instances consistently misclassified across models and runs could prove illuminating. These cases may reflect ambiguous or structurally atypical inputs that challenge current prompting paradigms. Understanding their distribution and linguistic characteristics may inform new strategies for model calibration, data augmentation, or annotation refinement.

\section*{Acknowledgement}
This work was supported by the HAMiSoN project grant CHIST-ERA-21-OSNEM-002, AEI PCI2022-135026-2 (MCIN/AEI/10.13039/501100011033 and EU “NextGenerationEU”/PRTR), DeepInfo (PID2021-127777OB-C22) project and UNED funding for open access publishing.
\bibliographystyle{elsarticle-num} 
\bibliography{bib_clean.bib}
\section*{Appendix A: Prompt Structure}
\label{appendix:prompt_templates}
To enhance clarity and interpretability, the prompts are colour-coded as follows:

\begin{itemize}
    \item \textcolor{introcolor}{\textbf{Blue}}: Instructional introduction or task formulation presented to the model.
    \item \textcolor{titlecolor}{\textbf{Green}}: Titles of narratives and sub-narratives.
    \item \textcolor{descriptioncolor}{\textbf{Grey}}: Descriptive content associated with each narrative or sub-narrative.
    \item \textcolor{guidecolor}{\textbf{Red}}: System-level instructions or control tokens, where applicable.
\end{itemize}
\newpage
{\footnotesize
{\color{introcolor}
\noindent Your primary role is to analyse a tweet \{tweet\} and categorise them according to predefined narrative themes that reflect different portrayals and perspectives of China.\\
Your classification should help in understanding the overarching sentiments and strategic messaging in public discourse.}

\par\vspace{10pt}\noindent

{\color{guidecolor}Narratives to Detect:}

\begin{itemize}
    \item[] \textcolor{titlecolor}{1: The West as Immoral and Hostile:}\\
    {\color{descriptioncolor}
    Identify tweets that depict Western countries, especially the US, as immoral, hostile, or decadent. Look for content where China is positioned as a victim of Western actions or policies.}

    \item[] \textcolor{titlecolor}{2: China as a Benevolent Power:}\\
    {\color{descriptioncolor}
    Classify tweets highlighting China's peaceful and cooperative international stance. Include tweets that mention China’s contributions to global peace, support for international law, and economic development in other nations.}

    \item[] \textcolor{titlecolor}{3: China’s Epic History:}\\
    {\color{descriptioncolor}
    Detect tweets that discuss China's historical resilience and achievements, particularly those crediting the Chinese Communist Party with overcoming adversities and leading national modernisation.}

    \item[] \textcolor{titlecolor}{4: China’s Political System and Values:}\\
    {\color{descriptioncolor}
    Look for tweets advocating for socialism with Chinese characteristics and portraying it as aligned with the will of the Chinese people. Tweets should suggest that China’s political system supports genuine democracy and global peace.}

    \item[] \textcolor{titlecolor}{5: Success of the Chinese Communist Party’s Government:}\\
    {\color{descriptioncolor}
    Identify tweets focusing on the government’s role in driving China’s technological, economic, and social advancements, such as achievements in 5G technology.}

    \item[] \textcolor{titlecolor}{6: China’s Cultural, Natural, and Heritage Appeal:}\\
    {\color{descriptioncolor}
    Classify tweets that promote Chinese culture, traditions, natural beauty, or heritage sites. This includes mentions of historical cities, cultural festivities, and natural landscapes.}
\end{itemize}

\par\vspace{10pt}\noindent

{\color{guidecolor}
Instructions for Classification:
\begin{enumerate}
    \item Read carefully the tweet \{tweet\}
    \item Determine which narrative(s) it supports based on the content and sentiment expressed. A tweet may align with at most 2 narratives if it incorporates elements from more than one category. It is possible that the tweet does not support any narrative.
    \item You have to generate a JSON structure:
\end{enumerate}

\begin{verbatim}
{
"classification": [1, 2, 3, 4, 5, 6] at most 2 categories or [] 
if the tweet doesn't support any narrative,
"reasoning": "reasoning of the answer in maximum 50 words."
}
\end{verbatim}
}
}



\end{document}